%% file: main.tex
\documentclass[fleqn,11pt]{wlscirep}
\usepackage[utf8]{inputenc}
\usepackage[T1]{fontenc}
\usepackage{bm}
\usepackage{setspace}
\usepackage{float}
\usepackage{titlesec}
\usepackage{longtable}
\usepackage{tabularx}
\usepackage{array}
\usepackage{geometry}
\usepackage{multirow}
\usepackage{makecell,booktabs}
\usepackage{subcaption}
\newcolumntype{C}[1]{>{\rule{0pt}{0.4ex}\centering\arraybackslash}p{#1}}

\title{Genome-Anchored Foundation Model Embeddings Improve Molecular Prediction from Histology Images}

\author[1,$\ddagger$]{Cheng Jin}
\author[1,$\ddagger$]{Fengtao Zhou}
\author[2,3,4,5,$\ddagger$]{Yunfang Yu}
\author[1]{Jiabo Ma}
\author[1]{Yihui Wang}
\author[1]{Yingxue Xu}
\author[1]{Huajun Zhou}
\author[1]{Hao Jiang}
\author[6]{Luyang Luo}
\author[2]{Luhui Mao}
\author[2]{Zifan He}
\author[7]{Xiuming Zhang}
\author[7]{Jing Zhang}
\author[8]{Ronald Cheong Kin Chan}
\author[2,*]{Herui Yao}
\author[1,9,10,11,12,*]{Hao Chen}

\affil[1]{Department of Computer Science and Engineering, The Hong Kong University of Science and Technology, Hong Kong SAR, China}
\affil[2]{Guangdong Provincial Key Laboratory of Malignant Tumor Epigenetics and Gene Regulation, Guangdong-Hong Kong Joint Laboratory for RNA Medicine, Department of Medical Oncology, Sun Yat-sen Memorial Hospital, Sun Yat-sen University, Guangzhou, China}
\affil[3]{Guangdong Provincial Key Laboratory of Cancer Pathogenesis and Precision Diagnosis and Treatment, Joint Big Data Laboratory, Department of Medical Oncology, Shenshan Medical Center, Memorial Hospital of Sun Yat-sen University, Shanwei, China}
\affil[4]{Institute for AI in Medicine and faculty of Medicine, Macau University of Science and Technology, Taipa, Macao SAR, China}
\affil[5]{Department of Breast Surgery, The First Affiliated Hospital, Jinan University, Guangzhou, China}
\affil[6]{Department of Biomedical Informatics, Harvard University, Boston, USA}
\affil[7]{Department of Pathology, The First Affiliated Hospital, School of Medicine, Zhejiang University, Hangzhou, China}
\affil[8]{Department of Anatomical and Cellular Pathology, The Chinese University of Hong Kong, Hong Kong SAR, China}
\affil[9]{Department of Chemical and Biological Engineering, The Hong Kong University of Science and Technology, Hong Kong SAR, China}
\affil[10]{Division of Life Science, The Hong Kong University of Science and Technology, Hong Kong SAR, China}
\affil[11]{HKUST Shenzhen-Hong Kong Collaborative Innovation Research Institute, Shenzhen, China}
\affil[12]{State Key Laboratory of Nervous System Disorders, The Hong Kong University of Science and Technology, Hong Kong SAR, China}

\affil[ ]{$^\ddagger$ \textit{Contributed equally}}
\affil[ ]{$^*$ \textit{\textbf{Corresponding Author}}: Hao Chen (jhc@cse.ust.hk) and Herui Yao (yaoherui@mail.sysu.edu.cn)}

\begin{abstract}
Precision oncology requires accurate molecular insights, yet obtaining these directly from genomics is costly and time-consuming for broad clinical use. Predicting complex molecular features and patient prognosis directly from routine whole-slide images (WSI) remains a major challenge for current deep learning methods. Here we introduce PathLUPI, which uses transcriptomic privileged information during training to extract genome-anchored histological embeddings, enabling effective molecular prediction using only WSIs at inference. Through extensive evaluation across 49 molecular oncology tasks using 11,257 cases among 20 cohorts, PathLUPI demonstrated superior performance compared to conventional methods trained solely on WSIs. Crucially, it achieves AUC $\geq$ 0.80 in 14 of the biomarker prediction and molecular subtyping tasks and C-index $\geq$ 0.70 in survival cohorts of 5 major cancer types. Moreover, PathLUPI embeddings reveal distinct cellular morphological signatures associated with specific genotypes and related biological pathways within WSIs. By effectively encoding molecular context to refine WSI representations, PathLUPI overcomes a key limitation of existing models and offers a novel strategy to bridge molecular insights with routine pathology workflows for wider clinical application.
\end{abstract}

\begin{document}

\flushbottom
\maketitle

\thispagestyle{empty}
\setlength{\parskip}{0.8\baselineskip}
\input{Sections/Introduction}
\input{Sections/Results}
\input{Sections/Discussions}
\input{Sections/Method}

\noindent\scalebox{1.2}{\textbf{Data Availability}}\\
WSIs and corresponding transcriptomics profile from TCGA cohort were obtained from the public Genomic Data Commons (GDC) portal (\url{https://portal.gdc.cancer.gov}). Similarly, WSIs from the Clinical Proteomic Tumor Analysis Consortium (CPTAC) cohort were retrieved through the same platform. Private institutional cohort WSIs were obtained under data transfer agreements with collaborating hospitals. These datasets are not publicly available due to patient privacy obligations, institutional review board requirements, and data use agreements. However, researchers interested in accessing de-identified data may submit a reasonable request directly to the corresponding authors, subject to obtaining the necessary ethical approvals and complying with institutional policies. 

\noindent\scalebox{1.2}{\textbf{Code Availability}}\\
All model source code is available under a CC BY-NC-ND 4.0 license at a GitHub repository.

\noindent\scalebox{1.2}{\textbf{Acknowledgement}}\\
This work was supported by the National Natural Science Foundation of China (No. 62202403), Innovation and Technology Commission (Project No. MHP/002/22 and ITCPD/17-9), Research Grants Council of the Hong Kong Special Administrative Region, China (Project No: T45-401/22-N) and National Key R\&D Program of China (Project No. 2023YFE0204000).

\noindent\scalebox{1.2}{\textbf{Author contributions}}\\
C.J. and F.Z. investigated and designed the study. C.J., F.Z., and J.M. curated the public cohort dataset. Y.Y., L.Y., Z.H., J.M., Y.W., Y.X., H.Z., and X. Z. curated the private cohort dataset. Data quality control and preprocessing were completed by X.Z., Y.Y. and F.Z. C.J. and F.Z. designed and implemented the PathLUPI algorithm. C.J. conducted validation experiments with baseline methods. Visualization and analysis of experiments were conducted by C.J. and F.Z. Clinical guidance was provided by R.C.K.C., X.Z., and H.Y. H.J., L.L., and H.C. reviewed and edited the manuscript. H.C. and H.Y. supervised this research.

\bibliography{sample}

\clearpage

\input{Sections/Supplementary}

\end{document}

%% file: Sections/Introduction.tex
\section*{Introduction}
\begin{figure}
    \centering
    \includegraphics[width=\linewidth]{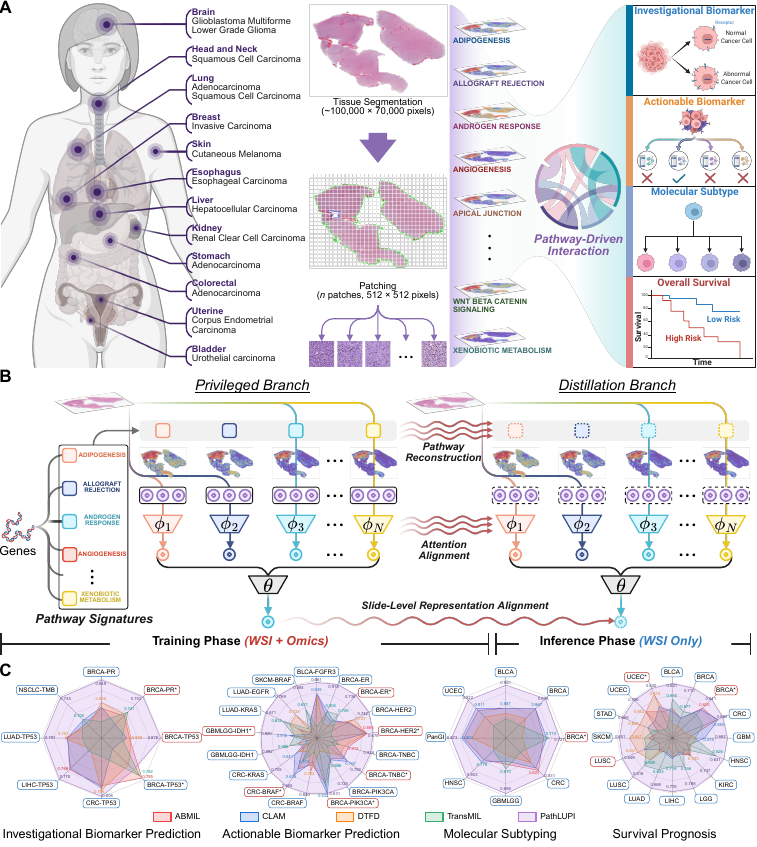}
    \caption{\textbf{Overview of the study.} \textbf{a.} Scope of PathLUPI framework development and validation phases, covering 13 distinct cancer types among essential molecular oncology tasks: Investigational biomarker prediction, actionable biomarker prediction, molecular subtyping, and survival prognosis. \textbf{b.} Schematic of PathLUPI paradigm. By leveraging the learning using privileged information (LUPI) paradigm, the model integrates transcriptomic pathway signatures as privileged supervisory signals during training to refine WSI representation learning. This process enables molecular-aware predictions using WSI images alone during inference. \textbf{c.} Performance comparison of PathLUPI against four representative frameworks (ABMIL, CLAM, DTFD, TransMIL) across 49 molecular oncology tasks, including: Investigational biomarker prediction (\textit{n}=8), actionable biomarker prediction (\textit{n}=17), molecular subtyping (\textit{n}=8), and survival prognosis (\textit{n}=16). * indicates tasks with external validation cohorts.}
    \label{fig:overview}
\end{figure}

The integration of molecular profiling into oncology has revolutionized tumor characterization, enabling targeted therapies informed by genomic drivers \cite{zhong2021small, cheng2021nanomaterials}. However, the clinical implementation of molecularly targeted therapies remains constrained by the practical limitations of genomic testing, including cost barriers and analytical latency, which impede timely decision-making \cite{kather2018genomics}. Computational pathology (CPATH) addresses this challenge through deep learning-based analysis of hematoxylin and eosin (H\&E) stained whole-slide images (WSIs), aiming to infer molecular characteristics from routine histology \cite{vasudevan2022digital}. Illustrated in Figure \ref{fig:overview}a, this paradigm has emerged as a critical tool for bridging molecular diagnostics and clinical workflows, enabling the prediction of investigational biomarkers \cite{de2024androgen, suehnholz2024quantifying}, which are molecular features currently under active clinical investigation, as well as the identification of actionable biomarkers, defined as molecular changes with established therapeutic relevance \cite{van2021limited, fong2025multilineage}. It also facilitates molecular subtyping \cite{kather2019deep} and enables prognostic stratification through systematic analysis of tumor morphology \cite{amgad2024population}. These capabilities collectively advance the vision of morphology-driven precision oncology, where ubiquitous histology slides enable the prediction of molecular characteristics through computational analysis.

Despite the immense potential of CPATH, achieving reliable molecular inference for precision oncology hinges on accurately deciphering the often complex and subtle relationships between genotype and morphology within WSIs. Initial studies demonstrated feasibility in predicting certain key genetic alterations and prognostic markers \cite{kather2020pan, qu2021genetic, liang2023deep, volinsky2024prediction}, and the recent emergence of powerful pathology foundation models pre-trained on vast WSI datasets offers strong general feature representations \cite{chen2024towards, xu2024whole, lu2024visual}. However, significant challenges remain in translating these capabilities to predict more complex or nuanced molecular features directly from histology. For instance, predicting alterations in genes like \textit{BRAF, KRAS}, or \textit{PIK3CA} from WSIs has been shown to be more difficult \cite{baldus2010prevalence, qu2021genetic, niehues2023generalizable}, indicating inherent limitations in capturing complex molecular correlates when models learn exclusively from WSI data, even at scale, hindering the reliable linkage between histological phenotype and underlying genotype.

To overcome this fundamental limitation, new strategies are needed that can leverage molecular information to improve the learning of genotype-correlated histology features, without compromising the ultimate goal of WSI-based inference in clinical practice. The increasing availability of large-scale datasets pairing WSIs with rich molecular profiles offers a unique opportunity in this regard. We propose that incorporating molecular context during model training can significantly improve molecular inference from WSIs. Consequently, we developed PathLUPI, a framework inspired by the learning using privileged information (LUPI) paradigm \cite{vapnik2009new, lopez2015unifying} (Figure \ref{fig:overview}b). PathLUPI leverages transcriptomic profiles structured according to cancer hallmark pathways \cite{hanahan2011hallmarks} as privileged information exclusively during the training phase. This molecular knowledge guides the feature extraction process of the CONCH pathology foundation model \cite{chen2024towards, lu2024visual}, refining its pathology-specific feature representations to be more indicative of the underlying molecular landscape. Crucially, the resulting framework yields robust genome-anchored histological embeddings capable of significantly enhancing the prediction of complex molecular features using only WSIs during inference. This preserves the scalability and workflow advantages of conventional CPATH methods.

Through a pan-cancer analysis of 11,257 cases from 20 cohorts spanning 13 tumor types, we demonstrate how privileged genomic supervision resolves key limitations of models solely trained on WSIs (Figure~\ref{fig:overview}c). We evaluated PathLUPI across 49 diverse molecular oncology tasks, encompassing investigational biomarker prediction ($n$=8), actionable biomarker prediction ($n$=17), molecular subtyping ($n$=8), and survival prognosis ($n$=16). In internal evaluations using the TCGA dataset ($N$=6,427 cases, 13 cohorts), PathLUPI consistently achieved superior performance by extracting genome-anchored histological embeddings. Specifically, it significantly improved prediction accuracy for challenging biomarkers such as \textit{BRAF} and \textit{KRAS} (\textit{P} < 0.001) and demonstrated improved molecular subtyping capabilities compared to baseline methods. Furthermore, PathLUPI showed superior (\textit{P} < 0.001) prognostic performance (C-index) across these cohorts compared to baselines. Crucially, external validation using 4,830 cases from 7 independent cohorts confirmed PathLUPI's resilience to domain shift when assessed on a range of these molecular oncology tasks. Additionally, its embeddings enabled the discovery of interpretable morphological signatures linked to molecular drivers, highlighting its potential to bridge molecular insights with routine pathology for broader clinical application.

%% file: Sections/Results.tex
\section*{Results}
\subsection*{Study cohort characteristics}
To develop and evaluate PathLUPI, we utilized WSIs and matched bulk RNA-seq gene expression with corresponding clinical and molecular annotations from 13 cancer types (\textit{N}=6,427) available in The Cancer Genome Atlas (TCGA): bladder cancer (BLCA) \cite{cancer2014comprehensive}, breast cancer (BRCA) \cite{cancer2012comprehensive}, colorectal cancer (COAD/READ) \cite{guinney2015consensus, liu2018comparative}, esophageal cancer (ESCA) \cite{liu2018comparative}, head and neck squamous cell carcinoma (HNSC) \cite{cancer2015comprehensive}, kidney renal clear cell carcinoma (KIRC) \cite{cancer2015comprehensive_KIRC}, glioma (GBM/LGG) \cite{ceccarelli2016molecular}, hepatocellular carcinoma (LIHC) \cite{ally2017comprehensive}, lung adenocarcinoma (LUAD) \cite{cancer2014comprehensive_LUAD}, lung squamous cell carcinoma (LUSC) \cite{cancer2012comprehensive_LUSC}, gastric adenocarcinoma (STAD) \cite{liu2018comparative}, melanoma (SKCM) \cite{akbani2015genomic}, and uterine corpus endometrial carcinoma (UCEC) \cite{Levine2013-vr}. For external validation, we assembled both private and public cohorts, each comprising only WSIs. The external datasets included locally collected breast cancer WSIs from Center-1 (\textit{N}=2,045) and Center-2 (\textit{N}=1,527), obtained following approval by the respective institutional review boards, as well as publicly available cohorts from the Clinical Proteomic Tumor Analysis Consortium (CPTAC, \textit{N}=406)\cite{ellis2013connecting} and the EBRAINS Digital Tumor Atlas (\textit{N}=852)\cite{roetzer2022digital}. This dataset enabled a rigorous evaluation of PathLUPI across a broad range of prediction tasks. In total, we assessed 48 tasks spanning 20 cohorts, grouped into four categories: investigational biomarker prediction (\textit{n}=8), actionable biomarker prediction (\textit{n}=17), molecular subtyping (\textit{n}=8), and patient survival prognosis (\textit{n}=16). Each task addressed a clinically or biologically relevant scenario, ranging from genomic alterations to patient outcomes, providing a robust benchmark for assessing the generalizability and clinical utility of PathLUPI. Detailed task definitions and statistics are provided in the \textbf{Methods} section and Extended Data Tables \ref{tab:supp_tasks}–\ref{tab:targeted_biomarkers}.

\subsection*{Overview of the PathLUPI architecture}
PathLUPI learns genome-anchored embeddings for molecular prediction from histology via a dual-branch training framework under the LUPI paradigm \cite{vapnik2009new}, depicted in Figure~\ref{fig:overview}b.  The \textit{privileged branch} processes morphological features extracted from WSIs using CONCH \cite{lu2024visual} and paired, pathway-grouped transcriptomic features processed via multilayer perceptron (MLP). These features are fused using a cross-attention mechanism, with weights shared across both branches, enabling the learning of direct phenotype-genotype associations. Concurrently, the \textit{distillation branch} receives identical WSI features but lacks the corresponding transcriptomic profile, aiming to emulate the privileged branch. It first reconstructs WSI features into a corresponding simulated transcriptomic representation, which is then fused with the initial morphological features via the shared cross-attention mechanism. This shared architecture compels the distillation branch to simulate privileged multimodal context integration without direct transcriptomic access. Multilevel alignment losses, including reconstruction, attention alignment, and representation consistency, enforce effective knowledge transfer by driving the distillation branch's internal representations and attention patterns to approximate those of the privileged branch. Consequently, the distillation branch learns to generate genome-anchored embeddings from WSI input alone, representing morphological features imbued with learned molecular context. During inference, only the optimized distillation branch is activated for predictions. A detailed methodology is provided in the \textbf{Methods} section.

\subsection*{PathLUPI enhances biomarker prediction}
We first evaluated the capability of PathLUPI to predict clinically relevant molecular biomarkers directly from WSIs, testing both investigational biomarkers informing diagnosis and prognosis (e.g., \textit{TP53} and \textit{TMB} status) and actionable biomarkers guiding targeted therapies (e.g., \textit{BRAF}, \textit{FGFR3}, \textit{KRAS} status) \cite{havel2019evolving, juan2024development}. As baselines, we included ABMIL\cite{ilse2018attention}, CLAM\cite{lu2021data}, DTFD\cite{zhang2022dtfd}, and TransMIL\cite{shao2021transmil}, which are solely trained on WSIs and commonly employed as benchmarks in the literature \cite{kather2020pan, qu2021genetic, liang2023deep, volinsky2024prediction}. More details are provided in the \textbf{Methods} section.

\begin{figure}[h]
    \centering
    \includegraphics[width=\linewidth]{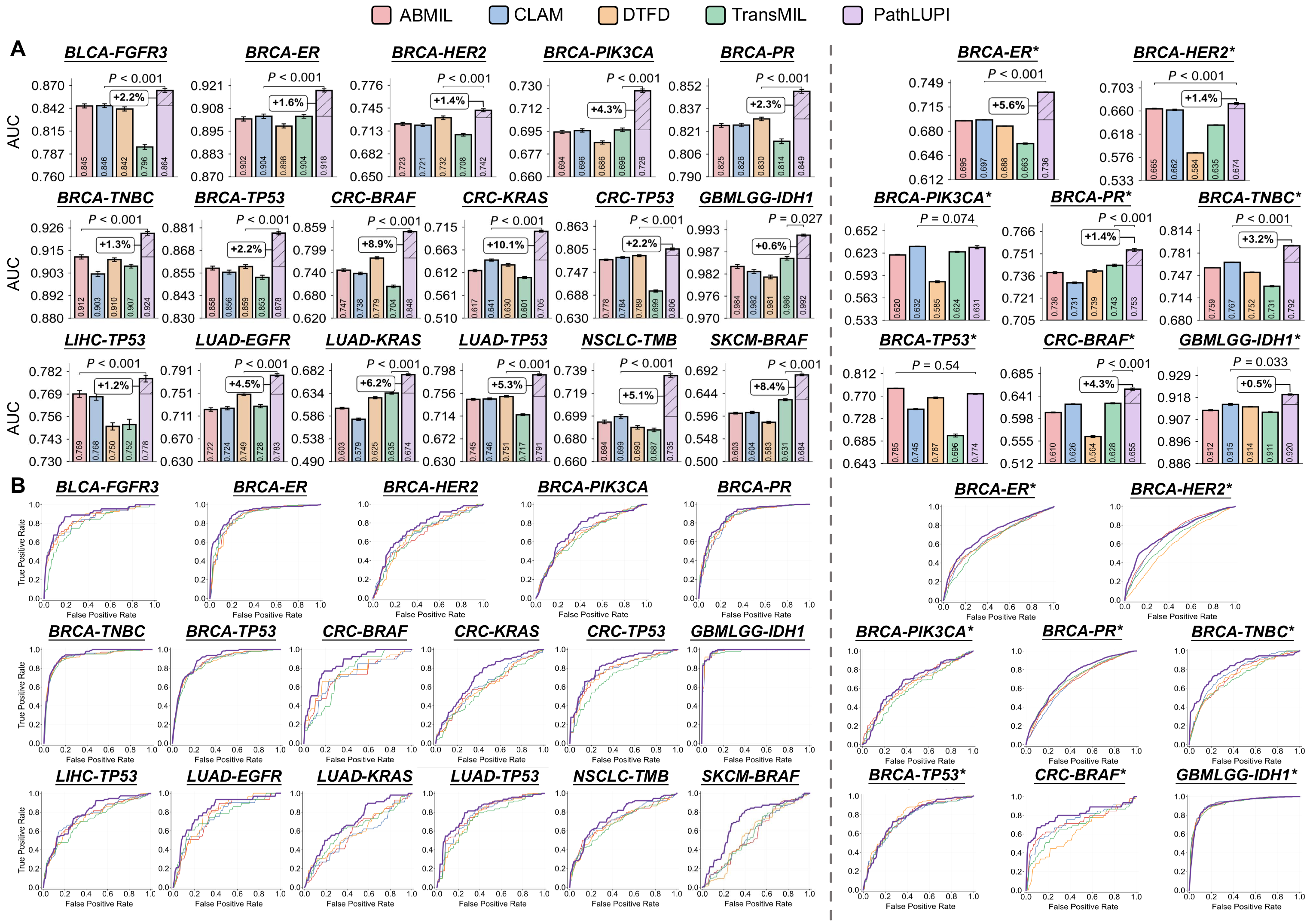} 
    
    \caption{Results of biomarker prediction across 25 tasks. \textbf{a.} Bar charts of mean AUC for each method. In each subfigure, the relative improvement of PathLUPI over the best baseline method and corresponding statistical significance is indicated. Error bars represent 95\% confidence intervals, and the centers correspond to the mean AUC values. \textbf{b.} Receiver operating characteristic (ROC) curves for each task. ROC curves were plotted by aggregating out-of-fold predictions from all 5 folds and applying 1,000 bootstrap resamplings. * denotes external cohorts. Detailed results are presented in Extended Data Tables \ref{tab:quantitative-1} and \ref{tab:quantitative-2}.}
    \label{fig:mut-pred}
\end{figure}

By integrating transcriptomic priors via the LUPI paradigm, PathLUPI significantly improves biomarker prediction accuracy. By integrating transcriptomic priors via the LUPI paradigm, PathLUPI significantly improves biomarker prediction accuracy. Illustrated in Figure \ref{fig:mut-pred}, among 25 prediction tasks, PathLUPI consistently outperformed the strongest baseline for each task. Averaged over all tasks, PathLUPI yielded a mean AUC increase of 4.48\% relative to the best-performing baseline. This improvement was observed in both internal validation (+4.83\%, \(P < 0.001\)) and external validation (+2.68\%, \(P < 0.001\)), indicating robust generalizability to independent cohorts. For actionable biomarkers, the mean AUC increase was 5.11\%, while for investigational biomarkers, the increase was 2.78\%. PathLUPI achieved an AUC of at least 0.80 in over half of all tasks, and at least 0.90 in 20\% of tasks, with the highest frequencies observed among actionable biomarkers, highlighting its clinical potential.

Performance gains were especially pronounced for clinically actionable biomarkers. In the prediction of \textit{BRAF} mutation in colorectal cancer, PathLUPI achieved an internal AUC of 0.840, corresponding to a 8.9\% increase over DTFD, and an external AUC of 0.655, a 4.3\% improvement over CLAM (\(P < 0.001\)). For \textit{KRAS} mutations, the internal AUC was 0.705, representing a 10.1\% increase relative to CLAM (\(P < 0.001\)). For \textit{PIK3CA} mutations in breast cancer, PathLUPI attained an internal AUC of 0.726, a 4.3\% improvement over TransMIL (\(P < 0.001\)), while the external AUC of 0.631 was comparable to the best baseline. In addition, for \textit{IDH1} mutation status in gliomas, PathLUPI achieved near-perfect accuracy, with an internal validation AUC of 0.992. In external validation, its AUC of 0.920 was comparable to that of the second-best method, CLAM. Overall, PathLUPI matched or outperformed the best baseline models trained solely on WSIs across nearly all tasks, with only minor exceptions in external \textit{PIK3CA} and \textit{TP53} mutation prediction. Comprehensive results and baseline comparisons are summarized in Extended Data Tables \ref{tab:quantitative-1} and \ref{tab:quantitative-2}.

\subsection*{PathLUPI improves molecular subtyping}
We further assessed the utility of PathLUPI for molecular subtyping, a critical task in precision oncology that classifies tumors into biologically distinct groups based on their molecular profiles to inform prognosis and guide therapeutic decisions \cite{dawson2013analysis, mosse2021pan}. Rather than isolated molecular events, it involves complex, multi-omics signatures, requiring models to detect subtle histological patterns and integrate them to reflect system-level transcriptomic changes. Addressing the need for comprehensive benchmarks, our study presents the most extensive evaluation of WSI-based molecular subtyping to date, systematically assessing performance across seven distinct cancer types. These include both organ-specific tasks and multi-type subtyping tasks, such as colorectal cancer (CRC) \cite{cancer2012comprehensive}, glioma subtypes (GBMLGG) \cite{ceccarelli2016molecular}, and a pan-gastrointestinal (PanGI) task \cite{liu2018comparative} that unifies several gastrointestinal malignancies under a single subtyping framework.

\begin{figure}[h]
    \centering
    \includegraphics[width=\linewidth]{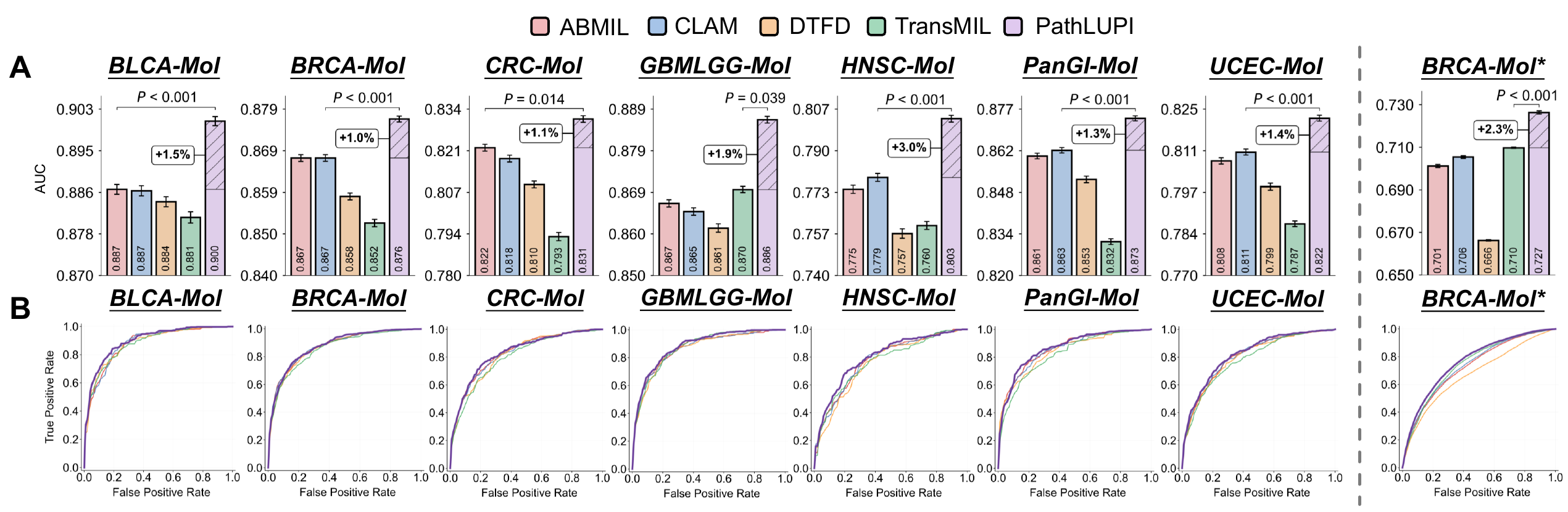} 
    
    \caption{Results of molecular subtyping on 8 tasks. \textbf{a.} Bar charts of mean AUC for each method. In each subfigure, the relative improvement of PathLUPI over the best baseline method and corresponding statistical significance is indicated. Error bars represent 95\% confidence intervals, and the centers correspond to the mean AUC values. \textbf{b.} Receiver operating characteristic (ROC) curves for each task. ROC curves were plotted by aggregating out-of-fold predictions from all 5 folds and applying 1,000 bootstrap resamplings. * denotes external cohorts. Detailed results are presented in Extended Data Table \ref{tab:quantitative-3}.}
    \label{fig:mol-sub}
\end{figure}

Despite the inherent difficulty of this task, PathLUPI achieved the highest overall mean AUC in both internal and external validations, as illustrated in Figure \ref{fig:mol-sub}. In internal validation, PathLUPI reached a mean AUC of 0.856, representing a 1.47\% improvement over the best-performing baseline ($P<0.001$). In external validation, PathLUPI maintained its advantage with a mean AUC of 0.727, surpassing the strongest comparator by 2.4\% ($P<0.001$). Although the absolute performance gains were modest compared to those observed in biomarker prediction, these results reflect the high level of difficulty associated with molecular subtyping from WSI data. Importantly, PathLUPI consistently outperformed all baselines across cancer types and subtyping tasks, demonstrating robust generalizability. Detailed results are available in Extended Data Table \ref{tab:quantitative-3}.

When applied to the more challenging task of molecular subtyping, PathLUPI again demonstrated the robustness observed during biomarker prediction. These findings reinforce that the integration of privileged structured transcriptomic supervision via the LUPI paradigm provides a reproducible advantage to achieve the stability and generalizability crucial for reliable molecular subtyping.

\subsection*{PathLUPI advances survival prognosis}
\begin{figure}[h]
    \centering
    \includegraphics[width=\linewidth]{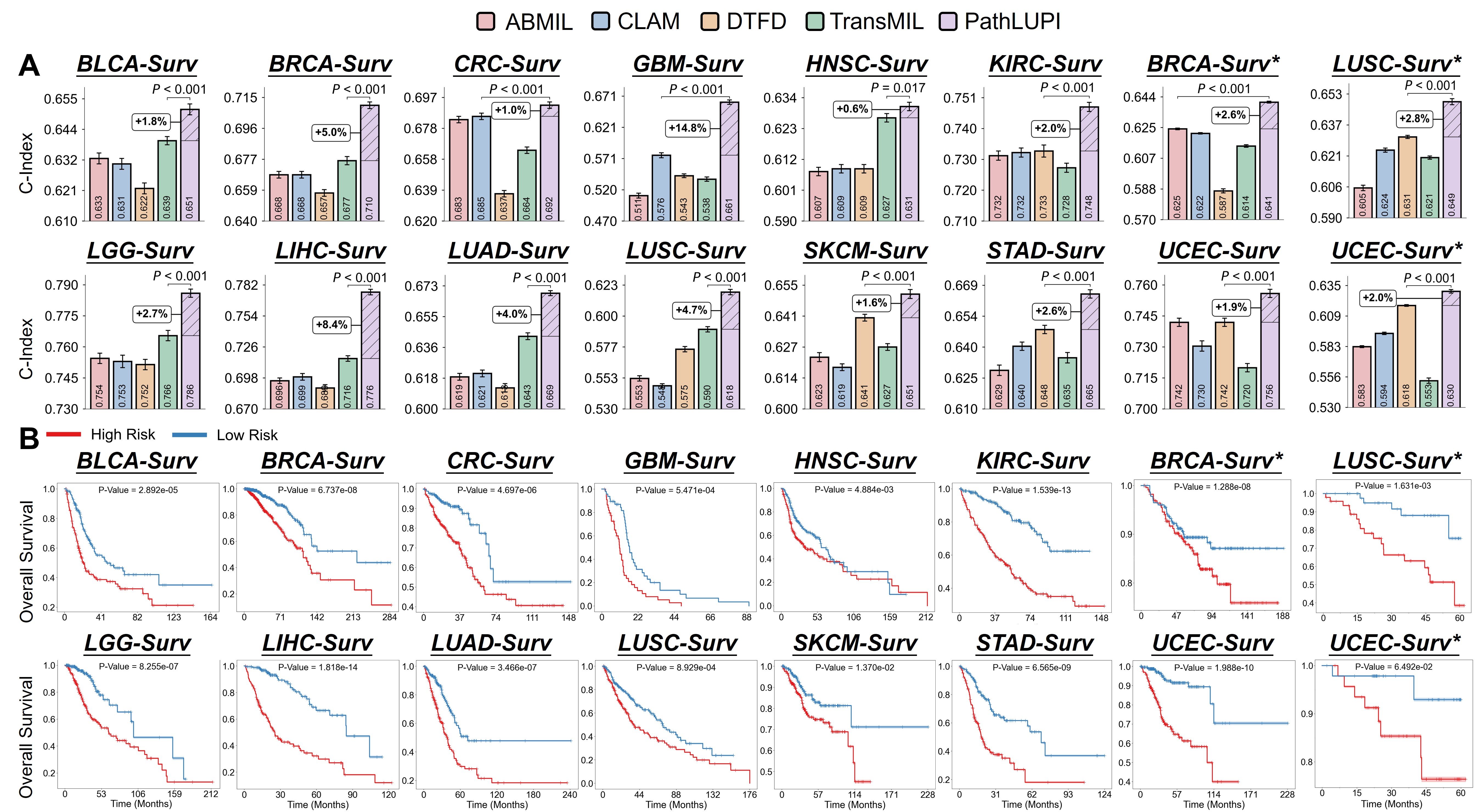}
    \caption{Results of survival prognosis on 16 tasks. \textbf{a.} Bar charts of mean AUC for each method. In each subfigure, the relative improvement of PathLUPI over the best baseline method and corresponding statistical significance is indicated. Error bars represent 95\% confidence intervals, and the centers correspond to the mean AUC values.
    \textbf{b.} Kaplan-Meier survival curves for high-risk and low-risk groups as predicted by PathLUPI. * denotes external cohorts. Detailed results are presented in Extended Data Table \ref{tab:quantitative-4}.}
    \label{fig:quant-survival}
\end{figure}

To further evaluate the clinical relevance of PathLUPI's genome-anchored embeddings, we assessed its performance in survival prognosis, a vital component of oncological assessment and treatment planning. We benchmarked PathLUPI against established baselines across a broad pan-cancer setting, encompassing 12 distinct cancer types for internal validation and 3 independent external validation cohorts. Evaluation across these diverse cohorts revealed PathLUPI's strength in survival prognosis, marked by the highest average concordance index (C-index) overall, as depicted in Figure \ref{fig:quant-survival}. revealed PathLUPI's strength in survival prognosis, marked by the highest average concordance index (C-index) overall. Among the baselines, TransMIL consistently emerged as the strongest comparator in internal validation, while CLAM achieved the best performance in external cohorts.

In internal validation, PathLUPI achieved a mean C-index of 0.693, representing a 5.19\% improvement over TransMIL ($P<0.001$). Notably, PathLUPI exceeded the recognized clinical benchmark of 0.70 \cite{alba2017discrimination} in several major cancer types, including BRCA (0.710), LGG (0.786), LIHC (0.776), KIRC (0.747), and UCEC (0.756). The C-index of 0.661 in GBM represents a substantial 14.8\% improvement over the baseline, underscoring PathLUPI’s exceptional prognostic performance in this highly challenging cancer type. These results, together with consistent improvements across most cancer types, underscore the robust and generalizable nature of PathLUPI's prognostic capabilities. In external validation, PathLUPI maintained superior performance, achieving a mean C-index of 0.640 and surpassing the strongest baseline, CLAM (0.613, +4.35\%, $P<0.001$). Consistent improvements were observed in multiple cohorts, such as LUSC*, where PathLUPI outperformed DTFD (0.649, +2.84\%, $P<0.001$). These findings collectively demonstrate the resilience and broad applicability of PathLUPI across both internal and external datasets. A summary of results for all cancer types is provided in Extended Data Table \ref{tab:quantitative-3}.

These quantitative improvements signify enhanced patient stratification power, as visually supported by Kaplan-Meier analyses where PathLUPI achieved consistently clear and statistically significant separation between predicted high-risk and low-risk patient groups. Importantly, this strong prognostic ability was maintained in the three external validation cohorts, where PathLUPI again outperformed all comparators, demonstrating resilience to domain shifts. Together, these results demonstrate that PathLUPI’s integration of privileged transcriptomic supervision not only enhances quantitative performance but also delivers robust and generalizable patient stratification across diverse cancer types and clinical cohorts, supporting its translational value for future clinical deployment \cite{Lu2024NatProtoc, Wagner2024GenomeMed}.

\subsection*{PathLUPI reveals potential genetic-morphological links}
\begin{figure}[h]
    \centering
    \includegraphics[width=\linewidth]{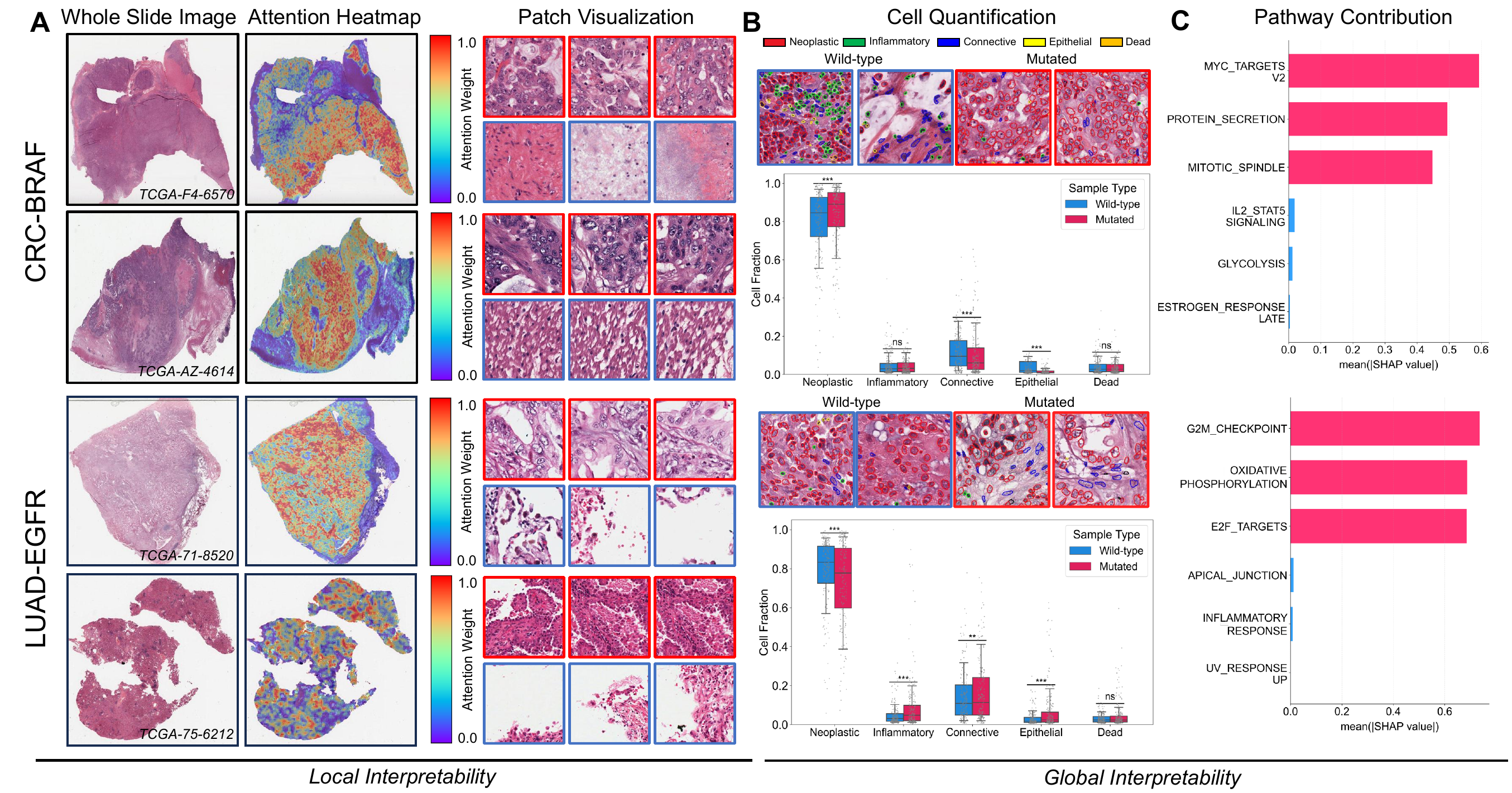}
    \caption{
\textbf{Local and global interpretability analyses of PathLUPI for two representative biomarker prediction tasks: colorectal cancer with \textit{BRAF} mutation and lung adenocarcinoma with \textit{EGFR} mutation.} 
\textbf{a.} Spatial attention heatmaps highlighting histological regions most relevant for molecular prediction in mutated samples. 
\textbf{b.} Quantification of cellular composition (neoplastic, inflammatory, epithelial, connective, dead cells) in the top 1\% high-attention patches across the exemplar cohorts. 
\textbf{c.} Shapley value attribution of the top-3 and bottom-3 transcriptomic pathways influencing model predictions in each cohort. Results for other tasks can be found in the Extended Data Figures \ref{fig:supp-1}-\ref{fig:supp-15}.
}
    \label{fig:qualitative}
\end{figure}

Beyond quantitative improvements in prediction tasks, PathLUPI provides an opportunity to explore and visualize the spatial relationships between molecular alterations and histopathological features. To achieve this, we systematically analyzed the cross-attention weights learned by the model, which quantify the contribution of each histological image patch to the prediction of specific transcriptomic pathways. Here, we define \textit{local interpretability} as the explanation of model predictions at the level of individual WSIs, revealing which tissue regions are most relevant for a specific task object. In contrast, \textit{global interpretability} refers to cohort-level analysis, identifying recurrent morphological patterns and cellular compositions associated with molecular features across multiple cases. We performed both local and global interpretability analyses for all internal biomarker prediction tasks. Figure~\ref{fig:qualitative} presents detailed results for two representative biomarkers, \textit{BRAF} in colorectal cancer and \textit{EGFR} in lung adenocarcinoma, both of which are well-established driver mutations with distinct morphological implications\cite{leach2021oncogenic, roman2018kras}. While these examples illustrate clear genotype–phenotype associations, the morphological correlates of many other molecular alterations are not yet fully understood. PathLUPI provides a systematic approach to further investigate such relationships. Results for all other tasks are provided in Extended Data Figures \ref{fig:supp-1}–\ref{fig:supp-15}.

At the local level, we extracted the spatial attention weights from the distilled branch of PathLUPI for each WSI, quantifying the relative contribution of individual image patches to the prediction of specific molecular tasks. By mapping these normalized attention scores back onto the original tissue slides, we visualized the spatial distribution of genotype–phenotype associations, highlighting morphological regions most strongly implicated in the model’s molecular predictions. As shown in Figure ~\ref{fig:qualitative}a, these heatmaps often revealed that, in slides from patients with different molecular characteristics, the model’s attention was focused on tissue regions that display well-known morphological features linked to these factors. For example, in colorectal cancer cases with \textit{BRAF} mutations, the attention map highlighted areas with increased flat or sessile areas, mucinous histologic regions, and serrated glandular structures commonly seen in \textit{BRAF}-mutated tumors \cite{leach2021oncogenic}. Similarly, for lung adenocarcinoma with \textit{EGFR} mutations, the attention map tended to concentrate on regions showing micropapillary and lepidic growth patterns, reflecting histological changes commonly linked to \textit{EGFR} mutations \cite{russell2013correlation, coudray2018classification}. These findings indicate that PathLUPI’s attention maps not only show which regions are used for prediction, but also reflect real, recognizable histological changes associated with specific genetic alterations.

At the global level, we conducted two analyses to elucidate the morphological and functional basis of model predictions across the cohort.
First, we identified the top 1\% of highest-attention patches from all correctly predicted WSIs for each molecular task, and characterized their cellular composition using CellViT++\cite{horst2024cellvit, horst2025cellvit++}. This revealed characteristic distributions of tumor, inflammatory, epithelial, connective tissue, and dead cells associated with specific molecular features, as illustrated in Figure~\ref{fig:qualitative}b. For \textit{BRAF}-mutated colorectal cancers, high-attention regions showed marked enrichment of neoplastic cells, while wild-type samples exhibited higher proportions of connective tissue and epithelial cells ($P<0.001$). Inflammatory and dead cell fractions were comparable between groups. These observations are consistent with previous single-cell and spatial profiling studies, which have demonstrated increased tumor cellularity and distinct stromal characteristics in \textit{BRAF}-mutated tumors \cite{roerink2018intra, tian2023combined}. In contrast, for \textit{EGFR}-mutated lung adenocarcinoma, the cellular composition of high-attention regions showed a more complex pattern. While neoplastic cells remained predominant in both mutated and wild-type samples, \textit{EGFR}-mutated cases exhibited significantly higher proportions of inflammatory and connective tissue cells ($P < 0.01$), as well as a modest increase in epithelial cells ($P<0.001$). This unique cellular composition signature aligns with previous findings from single-cell RNA sequencing studies\cite{hanley2023single, han2024atlas}, which demonstrated that \textit{EGFR}-driven lung adenocarcinomas reshape their microenvironment by recruiting specific inflammatory cells and modifying stromal components. 
Second, by aggregating pathway-level attention scores across the cohort and applying a gradient-based Shapley value approximation \cite{sundararajan2017axiomatic}, we quantified the functional relevance of biological pathways, thereby highlighting the most influential pathways driving predictions for each molecular task and uncovering context-specific biological processes with clear morphological correlates. Illustrated in Figure~\ref{fig:qualitative}c, for colorectal cancers with \textit{BRAF} mutations, the model’s highest-attention regions were characterized by distinct histological features, and pathway analysis identified \textit{EPITHELIAL\_MESENCHYMAL\_TRANSITION}, \textit{MYC\_TARGETS\_V2}, and \textit{WNT\_BETA\_CATENIN\_SIGNALING} as the most influential pathways in \textit{BRAF}-mutant colorectal cancer, in line with previous studies demonstrating their involvement in tumor progression and molecular subtype specification\cite{paul2023non, zhong2024recurrent, klomp2024determining}. Similarly, in lung adenocarcinoma with \textit{EGFR} mutations, \textit{G2M\_CHECKPOINT}, \textit{OXIDATIVE\_PHOSPHORYLATION}, and \textit{E2F\_TARGETS} emerged as top contributors, corroborating established roles for these pathways in the oncogenic processes underlying \textit{EGFR}-driven lung cancer\cite{xu2020integrative, xiao2023heterogeneity}.

Collectively, these qualitative and quantitative interpretability analyses support the existence of robust genotype–phenotype correlations at the tissue level and demonstrate that PathLUPI not only delivers accurate predictions but also uncovers spatially and functionally meaningful links between molecular alterations and histological features. By enabling interpretable, spatially resolved visualization of these relationships, PathLUPI offers new opportunities for hypothesis generation and biological discovery in cancer research.

\subsection*{Validating the impact of foundation model embeddings in PathLUPI}
\begin{figure}[htbp]
    \centering
\includegraphics[width=0.85\linewidth]{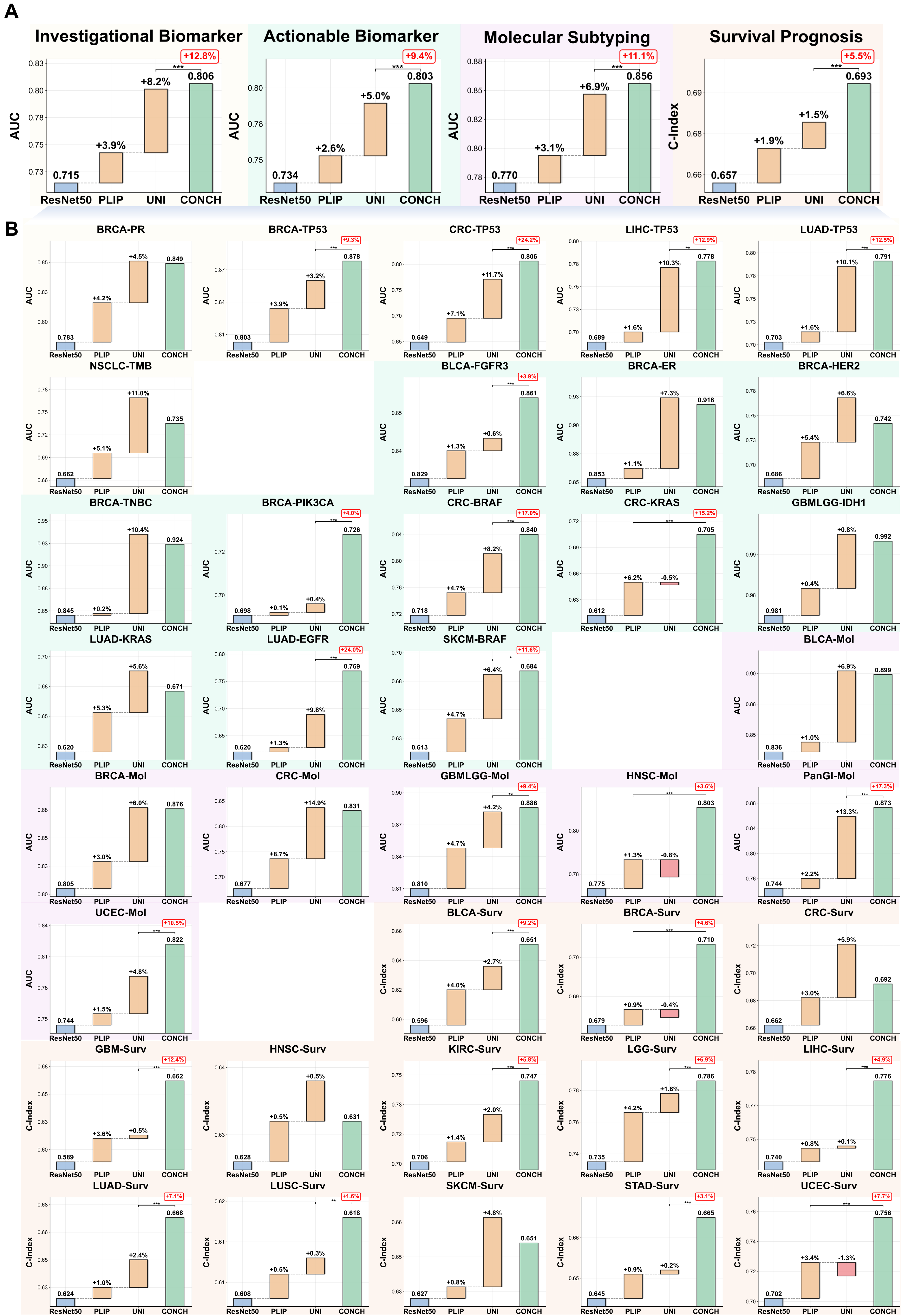}
    \caption{
Ablation study evaluating the impact of different foundation models on PathLUPI performance across all internal tasks.
\textbf{a.} Bridge plots summarizing model performance (AUC or C-index) for each task category using various feature extractors. The overall performance improvement from ResNet50 is indicated in the top-right corner of each subfigure.
\textbf{b.} Detailed performance comparison for individual tasks, with bars representing mean AUC or C-index across 5-fold cross-validation. Overall improvement and significance test results are displayed only when PathLUPI achieves the best performance in a given task.}

    \label{fig:ablation}
\end{figure}
To rigorously evaluate the impact of different foundation models for pathology embedding extraction within PathLUPI, we performed an ablation study in which the WSI feature extraction backbone was systematically replaced with alternative encoders. We compared three pathology-specific foundation models, PLIP\cite{huang2023visual}, UNI\cite{chen2024towards}, and CONCH\cite{lu2024visual}, and compared them with the widely used general-purpose ResNet50 baseline \cite{qu2021genetic,niehues2023generalizable}. Each configuration was independently trained and evaluated across four clinically relevant molecular oncology tasks, as summarized in Figure ~\ref{fig:ablation}a.

Across all evaluated tasks, all three pathology-specific foundation models consistently outperformed the ResNet50 baseline, with CONCH achieving the highest mean performance across tasks. Overall, for investigational biomarker prediction, CONCH achieved a mean AUC of 0.806, representing an absolute improvement of 12.8\% over ResNet50. In actionable biomarker prediction, CONCH delivered a 9.4\% gain, while for molecular subtyping, the improvement was most pronounced, with CONCH increasing AUC by 11.7\%. In survival prognosis, CONCH achieved a C-index of 0.693, corresponding to a 5.5\% increase over the general-purpose baseline. Notably, both PLIP and UNI also provided consistent gains across all tasks, though CONCH demonstrated the highest performance throughout. However, detailed cohort-level comparisons depicted in pretraining strategies Figure ~\ref{fig:ablation}b revealed that no single foundation model was universally optimal. In several specific cohorts, such as BRCA-ER, NSCLC-TMB, CRC-Mol, and HNSC-Surv, UNI outperformed CONCH, highlighting the complementary strengths of different pathology-specific.

These findings establish the critical importance of pathology-specific pretraining for histopathological representation learning. Compared to the generic vision backbone ResNet, all domain-adapted foundation models deliver markedly superior predictive accuracy, robustness, and generalization across diverse clinical endpoints, confirming that domain adaptation is foundational for effective histological analysis. Notably, the systematic ablation reveals that both the strategy and scale of pretraining are crucial determinants of downstream performance. While PLIP benefits from multimodal supervision, its limited dataset constrains the expressiveness of its learned representations. In contrast, UNI, self-supervised trained on millions of pathology images, consistently surpasses PLIP, indicating that large-scale self-supervised visual pretraining can provide more robust and generalizable features than visual-language models constrained by smaller datasets. The highest performance is achieved by CONCH, which integrates both massive data scale and expert-level semantic grounding. This combination demonstrates that neither data scale nor semantic supervision alone is sufficient; their integration is essential for maximizing the representational capacity of computational pathology models. Collectively, these results underscore the need for continued innovation in multimodal and large-scale pretraining strategies to further enhance the connection between histological phenotypes and molecular genotypes, ultimately advancing the clinical translation of computational pathology.

%% file: Sections/Discussions.tex
\section*{Discussion}\label{sec3} 
In this study, we present PathLUPI, an LUPI framework that systematically integrates transcriptomic knowledge into the training process of histology models. While numerous deep learning models have been developed to predict molecular characteristics from routine H\&E whole-slide images, they often underutilize the paired molecular data available during training, typically treating transcriptomic profiles merely as prediction targets rather than as explicit sources of guidance. In contrast, PathLUPI anchors patch-level features to gene expression signals structured by biological pathways, thereby transforming standard H\&E WSIs into genome-aware embeddings that preserve both visual and molecular information. Comprehensive evaluation across 49 benchmark tasks, encompassing 13 cancer types and 20 cohorts with a total of 11,257 patients, demonstrates that these embeddings consistently and substantially outperform traditional WSI-only pipelines. Notably, PathLUPI enhances the prediction of driver mutations, immunohistology markers, molecular subtypes, and survival risk. Furthermore, the pathway-based structure introduced during training enables the identification of interpretable relationships between tissue morphology and underlying biology, thereby providing novel insights into spatial genotype–phenotype associations that are typically challenging to discern.

Addressing the central challenge of timely and scalable molecular profiling in oncology, our work aligns with a broader clinical need to reduce reliance on genomic sequencing. While next-generation sequencing remains the gold standard for comprehensive molecular characterization, it is often constrained by high cost, long turnaround times, and limited accessibility in many clinical settings. Biological advances such as liquid biopsy and circulating tumor cell (CTC) analysis have improved efficiency, but these methods still face issues of variable sensitivity, cancer-type specificity, and inconsistent performance across disease stages \cite{cristofanilli2004circulating, chemi2019pulmonary, lin2021circulating}. In contrast, computational pathology (CPATH) offers a practical and scalable solution by leveraging routinely available H\&E-stained slides for molecular inference. Unlike these biologically-based assays, CPATH can rapidly scale with increasing digital pathology data, and its performance is expected to improve as larger and more diverse datasets become available. To further advance the field, there is a growing need for frameworks that incorporate prior biological knowledge during model training, thereby guiding feature extraction toward histological patterns most indicative of the underlying molecular landscape. Our work demonstrates that the integration of pathway-level knowledge, such as the Hallmarks of Cancer \cite{hanahan2011hallmarks}, can substantially improve both predictive performance and interpretability. While Hallmark gene sets provide a robust starting point, more recent pathway resources such as Reactome \cite{milacic2024reactome}, KEGG \cite{kanehisa2025kegg}, and even data-driven gene sets may offer complementary or more context-specific representations. Continued exploration of these resources will further enhance the scalability and interpretability of molecular profiling in oncology.

In addition to improving interpretability and robustness through the use of biological priors, our pathway-informed framework demonstrates the capacity to capture spatially structured molecular variation embedded within histological patterns. This capability suggests that, when properly guided, histology-based models can approximate spatial molecular landscapes that are otherwise accessible only through spatial transcriptomics (ST). While ST provides high-resolution spatial molecular profiling, its widespread adoption remains constrained by practical factors discussed above. By leveraging global transcriptomic supervision, our framework enables the extraction of spatially coherent, genome-anchored histological features from routine H\&E slides. These representations can be readily used for spatial hypothesis generation, pre-screening of samples for downstream spatial profiling, and guiding spatially targeted research in settings where ST is not feasible. Thus, PathLUPI expands the toolkit for spatial molecular exploration in oncology, offering a scalable and accessible alternative for investigating spatial heterogeneity in clinical contexts. Although multimodal approaches remain the gold standard when comprehensive spatial and molecular data are available, PathLUPI fills a critical gap by enabling spatial molecular inference in settings where such data are lacking, thereby complementing and extending the reach of spatial biology in clinical oncology \cite{zhou2023cross, xu2023multimodal}.

Despite these advances, several challenges remain before clinical translation can be fully realized.  First, the framework's predictive power is intrinsically linked to the availability of extensive, high-quality paired WSI and molecular data, particularly transcriptomic profiles, during training \cite{cui2022harnessing}. The current scarcity of such large-scale multimodal resources presents a substantial obstacle for the field, widely recognized as a key challenge for multimodal machine learning in oncology \cite{cui2022harnessing, zhu2023multimodal}. Second, despite validation efforts across multiple external cohorts, achieving consistent performance across diverse clinical settings remains challenging due to domain shifts. These shifts arise from inherent variations in slide preparation, staining protocols, image acquisition across different centers, and potential dataset biases, highlighting the need to further enhance model generalizability \cite{cui2022harnessing}. Third, the implementation of this work utilizes transcriptomics as the sole privileged modality. The integration of additional molecular layers, such as proteomics, epigenomics, or metabolomics, may capture complementary aspects of tumor biology and further enrich the learned representations \cite{wayment2024towards}. Addressing these limitations will require concerted community efforts in multimodal data curation, the advancement of data-efficient learning strategies, and prospective validation in real-world clinical environments.

%% file: Sections/Method.tex
\section*{Methods}\label{sec4}

\subsection*{Patient cohorts and ethics}\label{sec4-1}
This study adhered to the Declaration of Helsinki and the International Ethical Guidelines for Biomedical Research Involving Human Subjects, with ethical approval granted by the Human and Artifact Research Ethics Committee of The Hong Kong University of Science and Technology (HREP-2024-0423). We analyzed anonymized WSIs and transcriptomic profiles from The Cancer Genome Atlas (TCGA), accessed via the Genomic Data Commons Portal (\url{https://portal.gdc.cancer.gov/}) and cBioPortal \cite{cerami2012cbio}, encompassing 13 prevalent solid cancer types: bladder cancer (BLCA) \cite{cancer2014comprehensive}, breast cancer (BRCA) \cite{cancer2012comprehensive}, colorectal cancer (COAD/READ) \cite{guinney2015consensus, liu2018comparative}, esophageal cancer (ESCA) \cite{liu2018comparative}, head and neck cancer (HNSC) \cite{cancer2015comprehensive}, renal cell carcinoma (KIRC) \cite{cancer2015comprehensive_KIRC}, glioma (GBM/LGG) \cite{ceccarelli2016molecular}, hepatocellular carcinoma (LIHC) \cite{ally2017comprehensive}, lung adenocarcinoma (LUAD) \cite{cancer2014comprehensive_LUAD}, lung squamous cell carcinoma (LUSC) \cite{cancer2012comprehensive_LUSC}, gastric cancer (STAD) \cite{liu2018comparative}, melanoma (SKCM) \cite{akbani2015genomic}, and uterine corpus endometrial carcinoma (UCEC) \cite{Levine2013-vr}. External validation incorporated both institutional and public datasets. Locally collected breast cancer specimens were obtained from Center-1 and Center-2 with approvals from their respective Research Ethics Boards. These were complemented by publicly available data from the Clinical Proteomic Tumor Analysis Consortium (CPTAC) \cite{ellis2013connecting} and the EBRAINS Digital Tumor Atlas \cite{roetzer2022digital}.

\subsection*{Molecular labels and clinical annotations}\label{sec4-2}
To evaluate the generalizability of the PathLUPI framework across diverse clinical and molecular prediction problems, we curated a set of 49 molecular oncology tasks spanning 13 cancer types. These tasks were defined based on biologically meaningful patient-level labels, including somatic mutations, immunohistochemistry (IHC) markers, molecular subtypes, and overall survival (OS). All internal tasks were constructed using TCGA data. Labels were matched to TCGA patient IDs, and for patients with multiple WSIs, slide-level features were aggregated into patient-level representations. Somatic mutation tasks were based on single-nucleotide variants (SNVs) and small insertions or deletions (INDELs), obtained from TCGA via the UCSC Xena platform \cite{goldman2020visualizing}. We identified 15 actionable biomarker tasks by cross-referencing mutated genes with FDA-approved targeted therapies (\url{https://www.cancer.gov/about-cancer/treatment/types/targeted-therapies/approved-drug-list}), and 9 investigational biomarker tasks based on frequently mutated genes lacking approved therapies, following prior work \cite{mendiratta2021cancer}. IHC-based labels (e.g., ER, PR, HER2, TNBC) were extracted from TCGA clinical annotations. Molecular subtype tasks were derived using TCGAbiolinks, including PAM50 status for BRCA \cite{cancer2012comprehensive}, and CMS status for colorectal cancer (COAD/READ) \cite{guinney2015consensus}. Survival prediction tasks used OS time and event status, available across all cancer types. Public external datasets were obtained from the CPTAC program via cBioPortal \cite{cerami2012cbio} and the institutional database maintained by EBRAINS \cite{roetzer2022digital}. Private external datasets were collected through collaborations with partner hospitals under appropriate data use agreements. All external datasets were processed using the same pipeline as TCGA to ensure consistency in feature extraction and label definition. A summary of all tasks, associated labels, and cancer-type coverage is provided in Extended Data Table \ref{tab:supp_tasks}.

\subsection*{Genome-anchored representation learning with privileged supervision}\label{sec4-3}

To enable biomarker prediction from WSIs while leveraging transcriptomic supervision during training, we adopt a dual-branch architecture under the learning using privileged information (LUPI) paradigm \cite{vapnik2009new, lopez2015unifying}.  
Each training case is denoted as \(\mathcal{C}_i = (W_i, \mathbf{g}_i)\), where \(W_i = \{W_i^{(1)}, W_i^{(2)}, \dots\}\) refers to one or more WSIs associated with case \(i\), and \(\mathbf{g}_i \in \mathbb{R}^d\) is the corresponding gene expression profile. Each WSI in \(W_i\) is independently processed into patch-level features, which are subsequently concatenated across all slides to form a unified image representation for the case.

\begingroup
\setlength{\parskip}{0.2em}  
\setlength{\parindent}{1em}  

\paragraph{Patch-level feature extraction and re-embedding.}
Each slide \(W_i^{(s)} \in W_i\) is divided into non-overlapping patches of size 512$\times$512 pixels at 20$\times$ magnification. We use the CONCH foundation model \cite{lu2024visual}, a pathology-specific pretrained encoder, to extract patch-level embeddings \(\mathbf{v}^{(s)}_j \in \mathbb{R}^{d_v}\), where \(j\) indexes the patches in slide \(W_i^{(s)}\), and \(d_v = 512\) denotes the patch embedding dimension. While these embeddings capture rich domain-specific information, they are derived in a task-agnostic manner and may not fully capture the fine-grained morphological cues that are potentially important for biomarker prediction. To mitigate this limitation, we incorporate a region-aware re-embedding transformer (RRT) module \cite{tang2024feature}, which aims to refine the original features by modelling local structures and spatial dependencies across tissue regions. Specifically, we partition each slide into \(R = 50\) latent spatial regions, corresponding to the number of transcriptomic pathways used in the subsequent step, to encourage the model to associate tissue patterns with biological processes. The re-embedded features retain the original dimensionality but carry richer contextual information. We denote the set of re-embedded patch features for slide \(i\) as \(\tilde{V}_i = [\tilde{\mathbf{v}}_1; \dots; \tilde{\mathbf{v}}_{N_i}] \in \mathbb{R}^{N_i \times d_v}\), where \(N_i\) is the number of patches in slide \(i\). These features serve as structured inputs for subsequent cross-modal alignment.

\paragraph{Transcriptomic embedding.} 
Each gene expression profile \(\mathbf{g}_i \in \mathbb{R}^d\) is a high-dimensional vector representing the transcriptomic state of case \(i\). To incorporate biological structure, we partition the gene vector into \(P = 50\) subsets based on the Hallmark gene sets from MSigDB \cite{liberzon2015molecular}, each corresponding to a distinct biological pathway. For each pathway \(p\), we extract the associated subvector \(\mathbf{g}_i^{(p)} \in \mathbb{R}^{G_p}\), and encode it using a dedicated multilayer perceptron (MLP) \(f_p: \mathbb{R}^{G_p} \rightarrow \mathbb{R}^{d_v}\). Since the number of genes \(G_p\) varies among pathways, each MLP is independently parameterized. The resulting embedding \(\mathbf{h}_i^{(p)} = f_p(\mathbf{g}_i^{(p)})\) provides a pathway-level feature vector in a shared latent space. Concatenating all pathway embeddings yields a structured transcriptomic representation \(\mathbf{Z}_i = [\mathbf{h}_i^{(1)}; \dots; \mathbf{h}_i^{(P)}] \in \mathbb{R}^{P \times d_v}\). This structured format facilitates fine-grained alignment with WSI-derived features in the privileged branch.

\paragraph{Privileged branch.}  
The privileged branch integrates transcriptomic signals into morphological representation learning by aligning image features with pathway-level molecular embeddings. To this end, we treat the transcriptomic matrix \(\mathbf{Z}_i \in \mathbb{R}^{P \times d_v}\) as a set of biologically informed queries that attend to the re-embedded patch feature \(\tilde{V}_i \in \mathbb{R}^{N_i \times d_v}\) through the cross-attention mechanism. This allows each pathway embedding to selectively extract morphological patterns from spatial regions associated with its biological function. Specifically, attention weights \(\boldsymbol{\alpha}_i^{\text{priv}} \in \mathbb{R}^{P \times N_i}\) are computed as \(\boldsymbol{\alpha}_i^{\text{priv}} = \text{softmax}((\mathbf{Z}_i \mathbf{W}_q)(\tilde{V}_i \mathbf{W}_k)^\top / \sqrt{d_k})\), where \(\mathbf{W}_q \in \mathbb{R}^{d_z \times d_k}\) and \(\mathbf{W}_k \in \mathbb{R}^{d_v \times d_k}\) are learnable projection matrices. The attended features are then obtained via \(\mathbf{F}_i^{\text{priv}} = \boldsymbol{\alpha}_i^{\text{priv}} (\tilde{V}_i \mathbf{W}_v)\), resulting in a pathway-aligned representation of shape \(\mathbb{R}^{P \times d_z}\). Finally, we apply gated attention pooling across pathways to produce a slide-level vector \(\mathbf{z}_i^{\text{priv}} \in \mathbb{R}^{d_z}\). Finally, we apply a learnable gated attention mechanism to aggregate pathway-level features into a slide-level representation \(\mathbf{z}_i^{\text{priv}}\), enabling the model to adaptively weigh biological pathways according to their morphological relevance for prediction.

\paragraph{Distilled branch.}  
To enable inference without transcriptomic input, the distilled branch learns to approximate the privileged representation using visual information alone. Instead of relying on measured gene expression, it derives a set of pseudo-pathway embeddings \(\hat{\mathbf{Z}}_i \in \mathbb{R}^{P \times d_v}\) from the re-embedded feature \(\tilde{V}_i \in \mathbb{R}^{N_i \times d_v}\), capturing morphological patterns that are predictive of pathway activity. These pseudo-embeddings serve as queries in the same cross-attention mechanism with shared parameters, producing attention weights \(\boldsymbol{\alpha}_i^{\text{distill}} \in \mathbb{R}^{P \times N_i}\) and attended features \(\mathbf{F}_i^{\text{distill}} \in \mathbb{R}^{P \times d_z}\). The same gated attention layer then aggregates these features into a slide-level vector \(\mathbf{z}_i^{\text{distill}} \in \mathbb{R}^{d_z}\), which is passed through the shared prediction head. This design enables the distilled branch to emulate pathway-aware inference in the absence of transcriptomics data.

\paragraph{Multi-level alignment and training objective.}  
To enable the distilled branch to approximate the privileged branch, we introduce auxiliary alignment objectives at three levels. At the pathway level, we apply a reconstruction loss that combines $\ell_1$ and soft cross-entropy terms, defined as \(\mathcal{L}_{\text{rec}} = \sum_{p=1}^{P} ( \| \hat{\mathbf{h}}_i^{(p)} - \mathbf{h}_i^{(p)} \|_1 + \text{SCE}(\hat{\mathbf{h}}_i^{(p)}, \mathbf{h}_i^{(p)}) )\). To promote consistency in spatial focus and global understanding, we additionally align the attention maps and slide-level representations of the two branches, resulting in losses \(\mathcal{L}_{\text{attn}}\) and \(\mathcal{L}_{\text{rep}}\), respectively. These alignment losses are further integrated with task-specific supervision to form the overall training objective. Each task is formulated as either a classification or a survival prediction problem, optimized via cross-entropy or negative log-likelihood with $\ell_1$ regularization, respectively. The supervised loss is defined as \(\mathcal{L}_{\text{sup}} = \mathcal{L}_{\text{sup}}^{\text{priv}} + \mathcal{L}_{\text{sup}}^{\text{distill}}\), where each term aggregates task-specific losses. The total loss is then given by \(\mathcal{L}_{\text{total}} = \mathcal{L}_{\text{sup}} + \lambda ( \mathcal{L}_{\text{rec}} + \mathcal{L}_{\text{attn}} + \mathcal{L}_{\text{rep}} )\), where \(\lambda\) controls the strength of auxiliary alignment.

\paragraph{Inference.}  
During inference, only the distilled branch is activated. Given a new case \(\mathcal{C}_i = (W_i, \cdot)\), where transcriptomic data is unavailable, patch-level features are extracted and re-embedded, followed by pseudo-pathway regression and cross-attention integration. The resulting representation \(\mathbf{z}_i^{\text{distill}}\) supports transcriptome-informed predictions directly from WSI.

\endgroup

\subsection*{Interpretability analysis}\label{sec4-4}

To evaluate the biological plausibility and clinical interpretability of the model’s predictions, we conducted interpretability analysis from two perspectives: \textit{local interpretability}, focusing on individual WSIs, and \textit{global interpretability}, capturing patterns across the patient cohort for each molecular oncology task.

\begingroup
\setlength{\parskip}{0.2em}  
\setlength{\parindent}{1em}  

\paragraph{Local interpretability.}
To visualize the spatially resolved genotype–phenotype associations captured by the model, we analyzed attention weights from the distilled branch of our framework. For each WSI, we retrieved the attention matrix \( \boldsymbol{\alpha}_i^{\mathrm{distill}} \in \mathbb{R}^{P \times N_i} \), where each element \( \alpha_i^{(p,j)} \) quantifies the relative contribution of patch \( j \) to pathway \( p \). Patch-level importance scores were then computed by averaging attention across all pathways as \( \bar{\boldsymbol{\alpha}}_i = \frac{1}{P} \sum_{p=1}^{P} \boldsymbol{\alpha}_i^{(p,:)} \in \mathbb{R}^{N_i} \). Since raw attention values are not directly comparable across samples, the resulting scores were transformed into percentiles ranging from 0 to 1, with higher values indicating stronger model attribution. These normalized scores were subsequently mapped back to the spatial coordinates of the original WSIs, enabling visualization of the inferred genotype–phenotype associations.

\paragraph{Global interpretability.}
To uncover cohort-level morphological and functional patterns underlying model predictions, we performed global interpretability analyses from two distinct perspectives. First, to characterize morphological features associated with high-confidence predictions, we identified the top 1\% highest-attention patches across all correctly predicted WSIs for each task. This proportion was chosen to ensure that selected patches are both abundant and representative of informative regions. We then analyzed the cellular composition of these patches using CellViT++ \cite{horst2024cellvit, horst2025cellvit++}, which classifies cells into five categories: tumor cells (red), inflammatory cells (green), connective tissue (blue), epithelial cells (yellow), and dead cells (orange). Second, to assess the functional relevance of each biological pathway at the cohort level, we analyzed the cross-attention matrix \( \boldsymbol{\alpha}_i^{\mathrm{distill}} \in \mathbb{R}^{P \times N_i} \) from the distilled branch of the framework. For each correctly predicted WSI, a pathway-level attention vector was computed by averaging over patches: \( \bar{\boldsymbol{\alpha}}_i = \frac{1}{N_i} \sum_{j=1}^{N_i} \boldsymbol{\alpha}_i^{(:,j)} \in \mathbb{R}^P \), where each element reflects the total attention received by a pathway from all patches. These vectors were aggregated across the cohort, and a gradient-based Shapley value approximation \cite{sundararajan2017axiomatic} was applied to the pathway-level attention vector of each WSI to estimate the individual contribution of each pathway to the model’s prediction. Finally, Shapley values were averaged across all correctly predicted WSIs within the cohort to obtain cohort-level pathway importance, which was visualized as bar plots to highlight the most and the least influential biological processes.
\endgroup

\subsection*{Implementation details}\label{sec4-5}
To benchmark the effectiveness of our PathLUPI framework, we compared it against several state-of-the-art baselines. These include ABMIL~\cite{ilse2018attention}, CLAM~\cite{lu2021data}, DTFD~\cite{zhang2022dtfd}, and TransMIL~\cite{shao2021transmil}, all of which rely solely on WSI for molecular oncology tasks, employing different variants of attention-based aggregation architectures. All models were implemented in PyTorch\cite{paszke2019pytorch}. WSI patching and attention heatmaps generation are through the CLAM toolbox \cite{lu2021data}. Cell-level quantification and analysis were conducted using CellViT++ \cite{horst2024cellvit, horst2025cellvit++}, a transformer-based framework specifically optimized for generalized cell segmentation and representation in histopathology. Model interpretability was assessed using SHAP (v0.46.0)\cite{lundberg2018explainable} to compute Shapley values. Data processing and visualization were conducted using NumPy (v1.24.1), pandas (v1.3.5), scikit-learn (v1.3.2), SciPy (v1.11.4), Matplotlib (v3.5.3), and Seaborn (v0.12.2). Each task-specific model was trained using five-fold cross-validation with an 80:20 train–validation split. For evaluation on external datasets, we applied each of the five cross-validation models independently and reported the average of their performance metrics, providing a more reliable estimate of model performance. Performance metrics were estimated using 1,000 bootstrap resamples to compute means and 95\% confidence intervals (CIs). Statistical significance between models was assessed using a one-sided Wilcoxon signed-rank test\cite{wilcoxon1992individual}, following prior work\cite{ma2024towards, xu2024multimodal}. All experiments were conducted on a single NVIDIA RTX 3090 GPU with 24 GB of memory. A complete list of model architectures and training details are provided in Extended Data Table \ref{tab:all_hyperparams}.

%% file: Sections/Supplementary.tex
\section*{Supplementary}
\captionsetup[figure]{labelformat=simple, labelsep=period, name=Extended Data Figure}
\captionsetup[table]{labelformat=simple, labelsep=period, name=Extended Data Table}

\setcounter{figure}{0}

\begin{figure}[h]
	\centering
	\includegraphics[width=\linewidth]{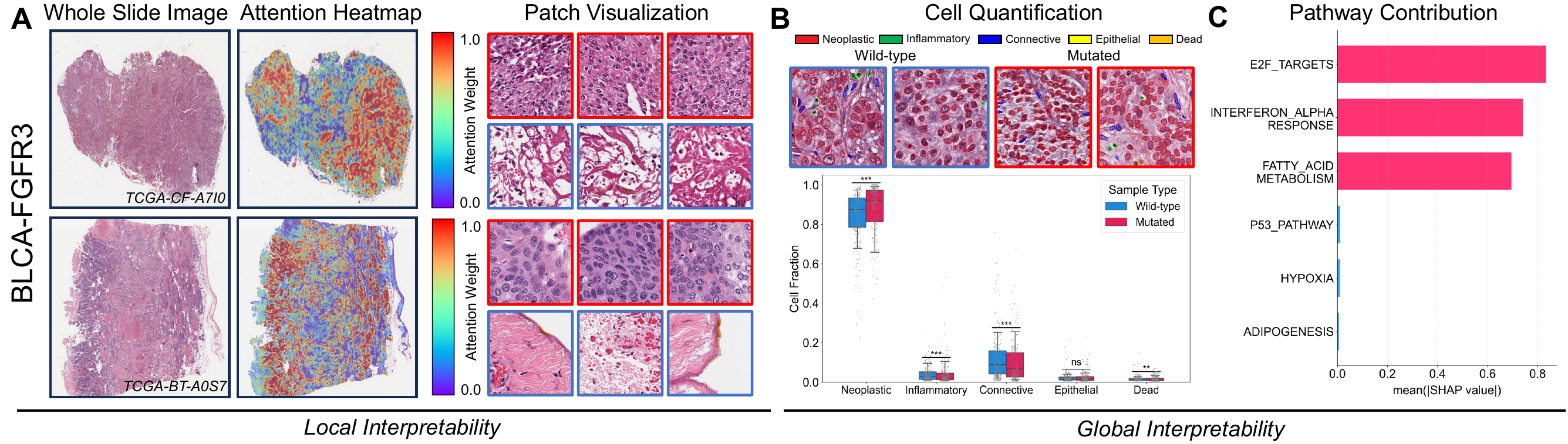}
	\caption{Local and global interpretability analyses of PathLUPI for bladder cancer with \textit{FGFR3} mutation status.
	\textbf{a.} Spatial attention heatmaps highlighting histological regions most relevant for molecular prediction in mutated samples. 
	\textbf{b.} Quantification of cellular composition (neoplastic, inflammatory, epithelial, connective, dead cells) in the top 1\% high-attention patches across the cohort. 
	\textbf{c.} Shapley value attribution of the top-3 and bottom-3 transcriptomic pathways influencing model predictions in each cohort.}
	\label{fig:supp-1}
\end{figure}

\begin{figure}[h]
	\centering
	\includegraphics[width=\linewidth]{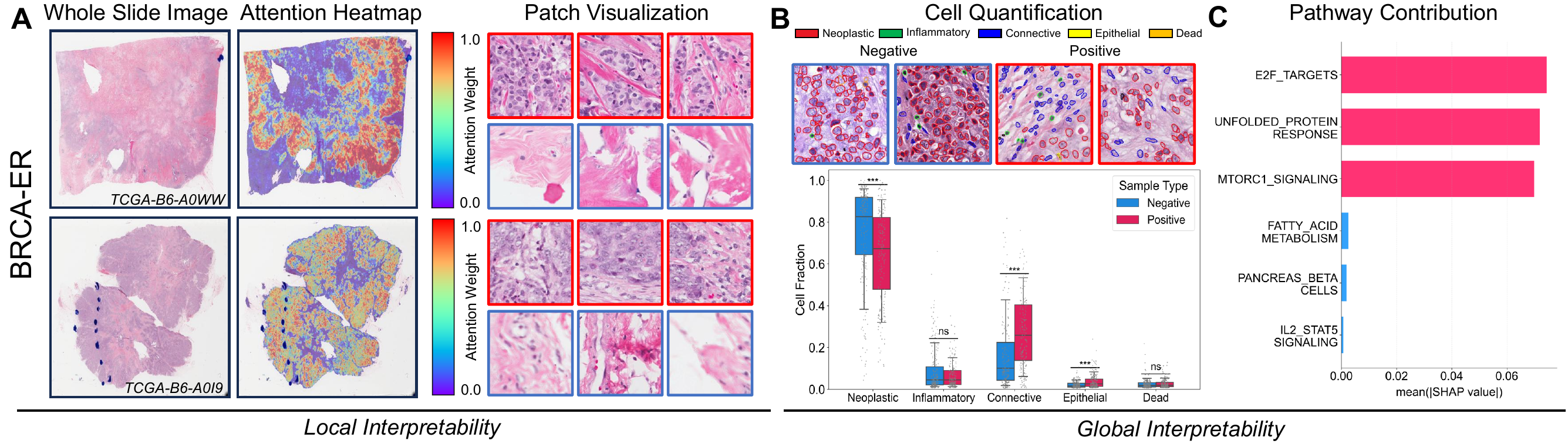}
	\caption{Local and global interpretability analyses of PathLUPI for breast cancer with estrogen receptor (ER) expression status. 
		\textbf{a.} Spatial attention heatmaps highlighting histological regions most relevant for molecular prediction in mutated samples. 
		\textbf{b.} Quantification of cellular composition (neoplastic, inflammatory, epithelial, connective, dead cells) in the top 1\% high-attention patches across the cohort. 
		\textbf{c.} Shapley value attribution of the top-3 and bottom-3 transcriptomic pathways influencing model predictions in each cohort.}
	\label{fig:supp-2}
\end{figure}

\begin{figure}[h]
	\centering
	\includegraphics[width=\linewidth]{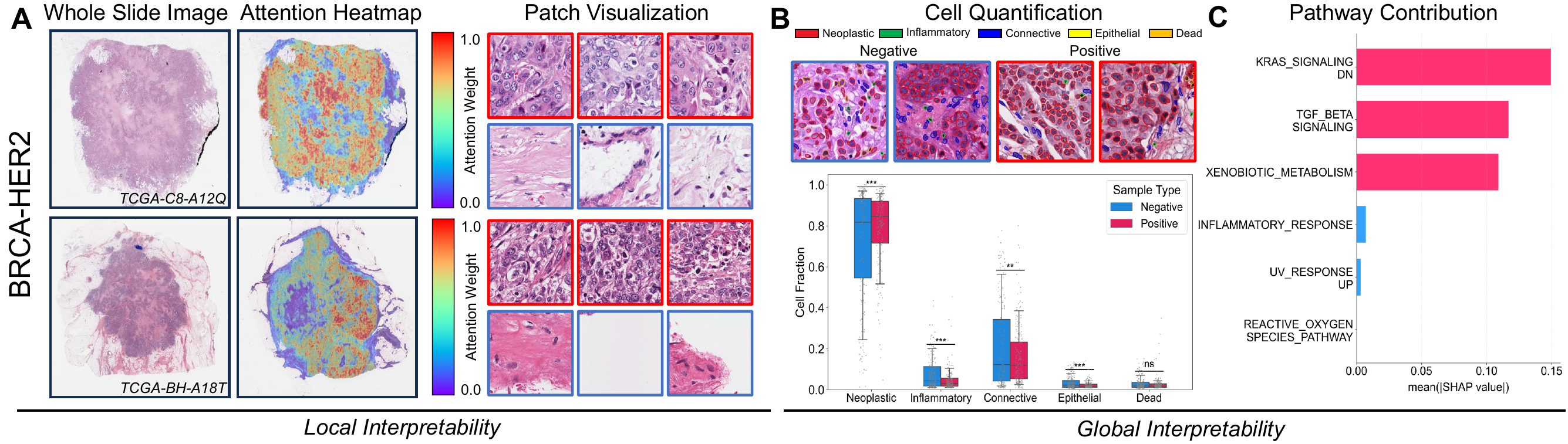}
	\caption{Local and global interpretability analyses of PathLUPI for breast cancer with human epidermal growth factor receptor 2 (HER2) expression status. 
		\textbf{a.} Spatial attention heatmaps highlighting histological regions most relevant for molecular prediction in mutated samples. 
		\textbf{b.} Quantification of cellular composition (neoplastic, inflammatory, epithelial, connective, dead cells) in the top 1\% high-attention patches across the cohort. 
		\textbf{c.} Shapley value attribution of the top-3 and bottom-3 transcriptomic pathways influencing model predictions in each cohort.}
	\label{fig:supp-3}
\end{figure}

\begin{figure}[h]
	\centering
	\includegraphics[width=\linewidth]{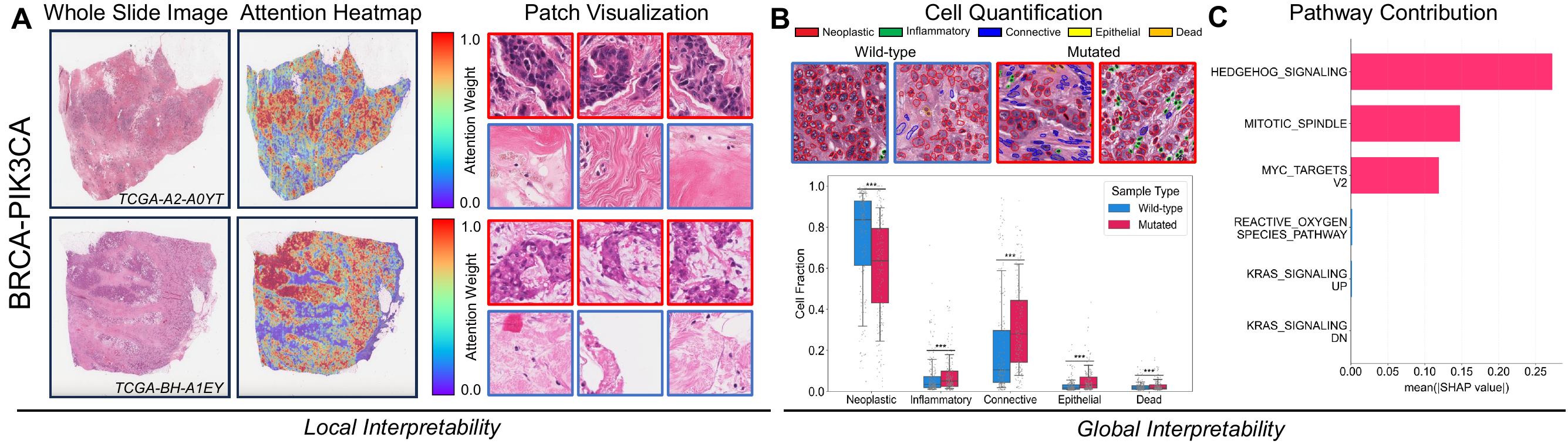}
	\caption{Local and global interpretability analyses of PathLUPI for breast cancer with \textit{PIK3CA} mutation status. 
		\textbf{a.} Spatial attention heatmaps highlighting histological regions most relevant for molecular prediction in mutated samples. 
		\textbf{b.} Quantification of cellular composition (neoplastic, inflammatory, epithelial, connective, dead cells) in the top 1\% high-attention patches across the cohort. 
		\textbf{c.} Shapley value attribution of the top-3 and bottom-3 transcriptomic pathways influencing model predictions in each cohort.}
	\label{fig:supp-4}
\end{figure}

\begin{figure}[h]
	\centering
	\includegraphics[width=\linewidth]{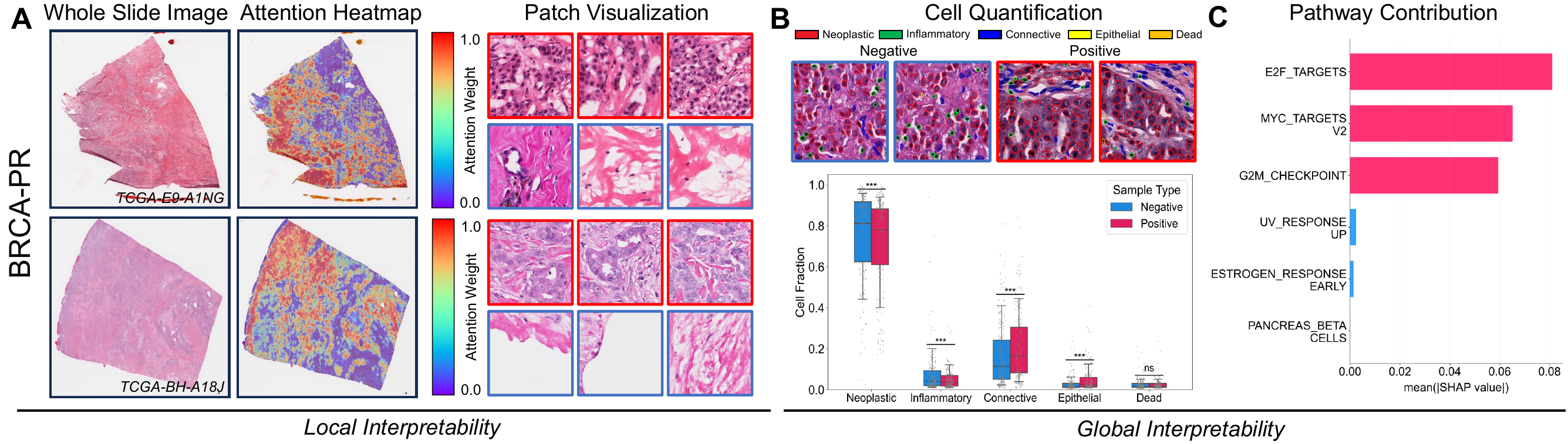}
	\caption{Local and global interpretability analyses of PathLUPI for breast cancer with progesterone receptor (PR) expression status. 
		\textbf{a.} Spatial attention heatmaps highlighting histological regions most relevant for molecular prediction in mutated samples. 
		\textbf{b.} Quantification of cellular composition (neoplastic, inflammatory, epithelial, connective, dead cells) in the top 1\% high-attention patches across the cohort. 
		\textbf{c.} Shapley value attribution of the top-3 and bottom-3 transcriptomic pathways influencing model predictions in each cohort.}
	\label{fig:supp-5}
\end{figure}

\begin{figure}[h]
	\centering
	\includegraphics[width=\linewidth]{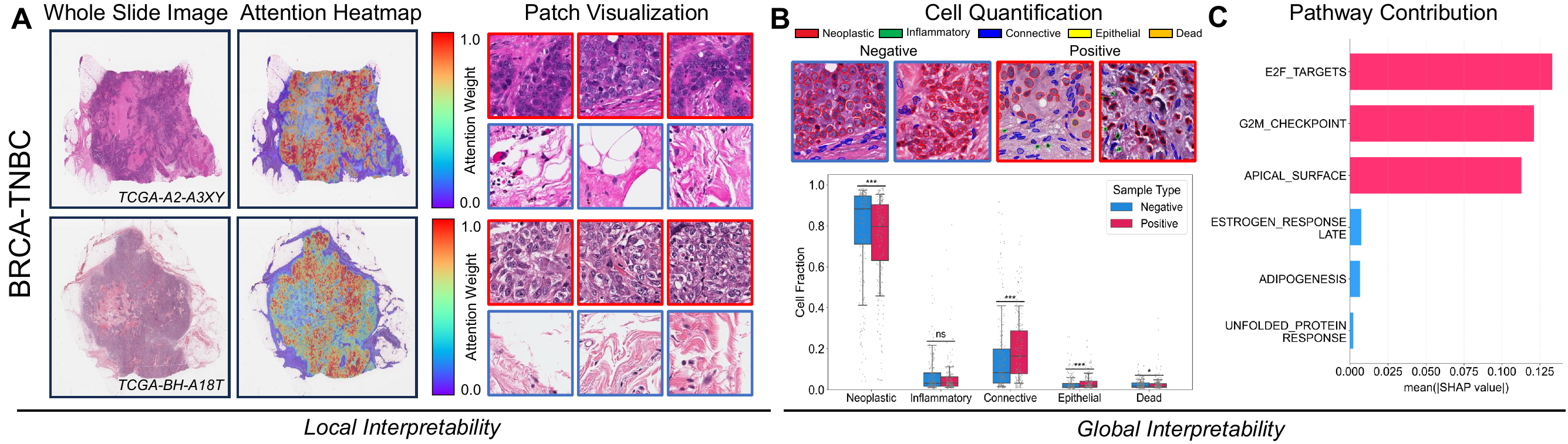}
	\caption{Local and global interpretability analyses of PathLUPI for triple-negative breast cancer (TNBC) status. 
		\textbf{a.} Spatial attention heatmaps highlighting histological regions most relevant for molecular prediction in mutated samples. 
		\textbf{b.} Quantification of cellular composition (neoplastic, inflammatory, epithelial, connective, dead cells) in the top 1\% high-attention patches across the cohort. 
		\textbf{c.} Shapley value attribution of the top-3 and bottom-3 transcriptomic pathways influencing model predictions in each cohort.}
	\label{fig:supp-6}
\end{figure}

\begin{figure}[h]
	\centering
	\includegraphics[width=\linewidth]{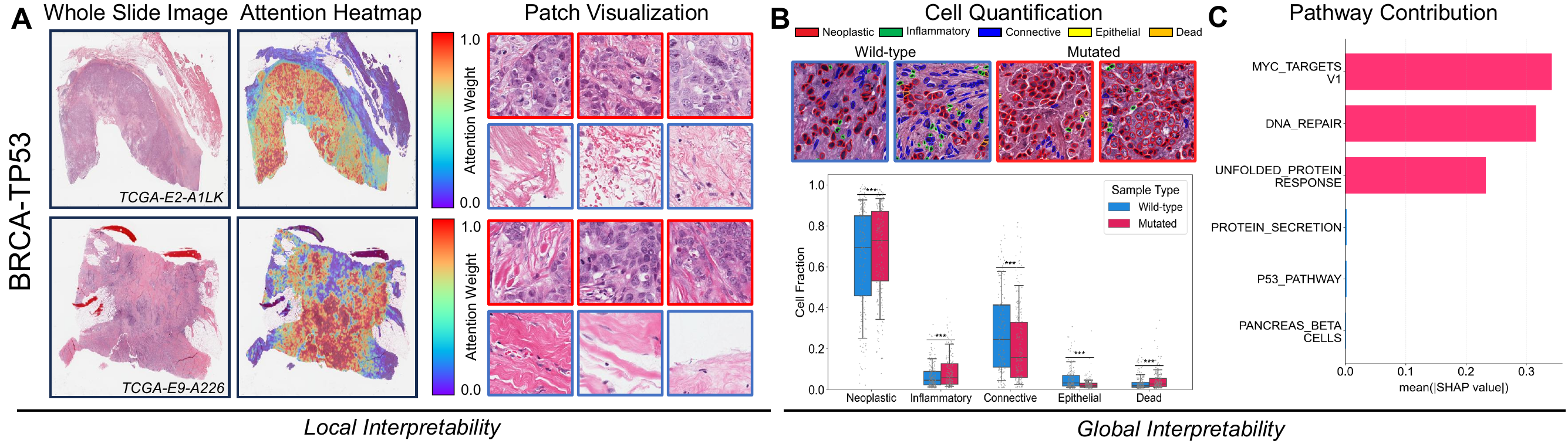}
	\caption{Local and global interpretability analyses of PathLUPI breast cancer with \textit{TP53} mutation status. 
		\textbf{a.} Spatial attention heatmaps highlighting histological regions most relevant for molecular prediction in mutated samples. 
		\textbf{b.} Quantification of cellular composition (neoplastic, inflammatory, epithelial, connective, dead cells) in the top 1\% high-attention patches across the cohort. 
		\textbf{c.} Shapley value attribution of the top-3 and bottom-3 transcriptomic pathways influencing model predictions in each cohort.}
	\label{fig:supp-7}
\end{figure}

\begin{figure}[h]
	\centering
	\includegraphics[width=\linewidth]{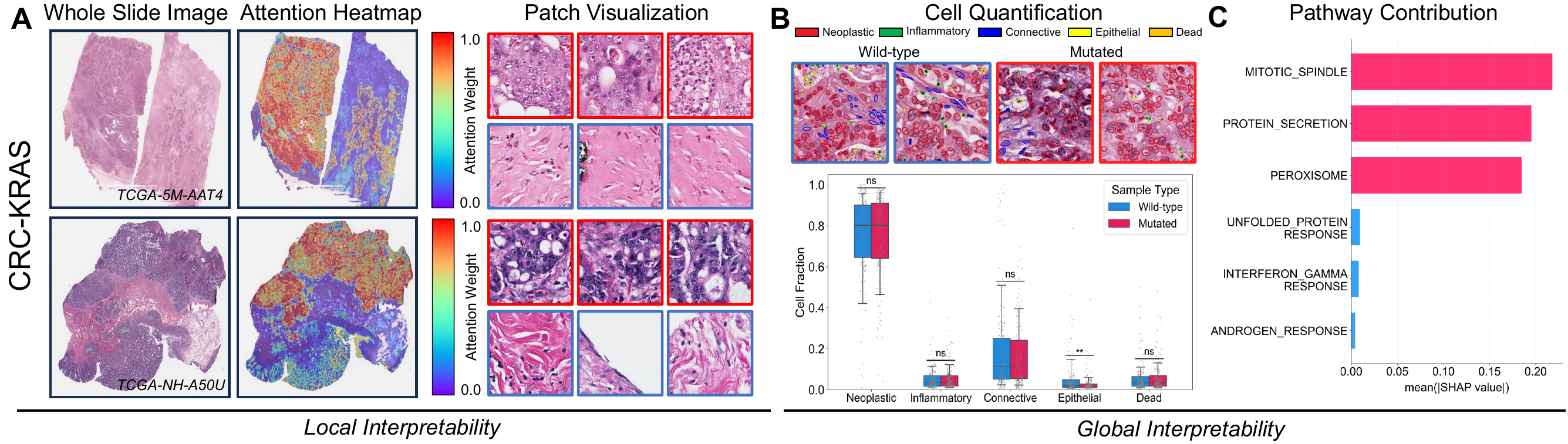}
	\caption{Local and global interpretability analyses of PathLUPI colorectal cancer with \textit{KRAS} mutation status. 
		\textbf{a.} Spatial attention heatmaps highlighting histological regions most relevant for molecular prediction in mutated samples. 
		\textbf{b.} Quantification of cellular composition (neoplastic, inflammatory, epithelial, connective, dead cells) in the top 1\% high-attention patches across the cohort. 
		\textbf{c.} Shapley value attribution of the top-3 and bottom-3 transcriptomic pathways influencing model predictions in each cohort.}
	\label{fig:supp-8}
\end{figure}

\begin{figure}[h]
	\centering
	\includegraphics[width=\linewidth]{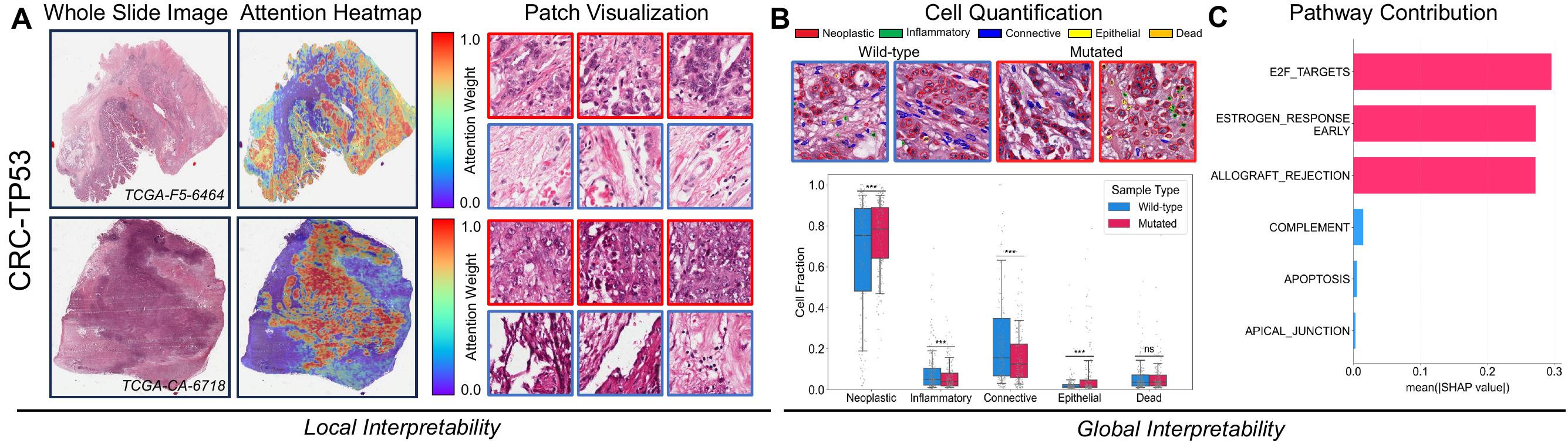}
	\caption{Local and global interpretability analyses of PathLUPI colorectal cancer with \textit{TP53} mutation status. 
		\textbf{a.} Spatial attention heatmaps highlighting histological regions most relevant for molecular prediction in mutated samples. 
		\textbf{b.} Quantification of cellular composition (neoplastic, inflammatory, epithelial, connective, dead cells) in the top 1\% high-attention patches across the cohort. 
		\textbf{c.} Shapley value attribution of the top-3 and bottom-3 transcriptomic pathways influencing model predictions in each cohort.}
	\label{fig:supp-9}
\end{figure}

\begin{figure}[h]
	\centering
	\includegraphics[width=\linewidth]{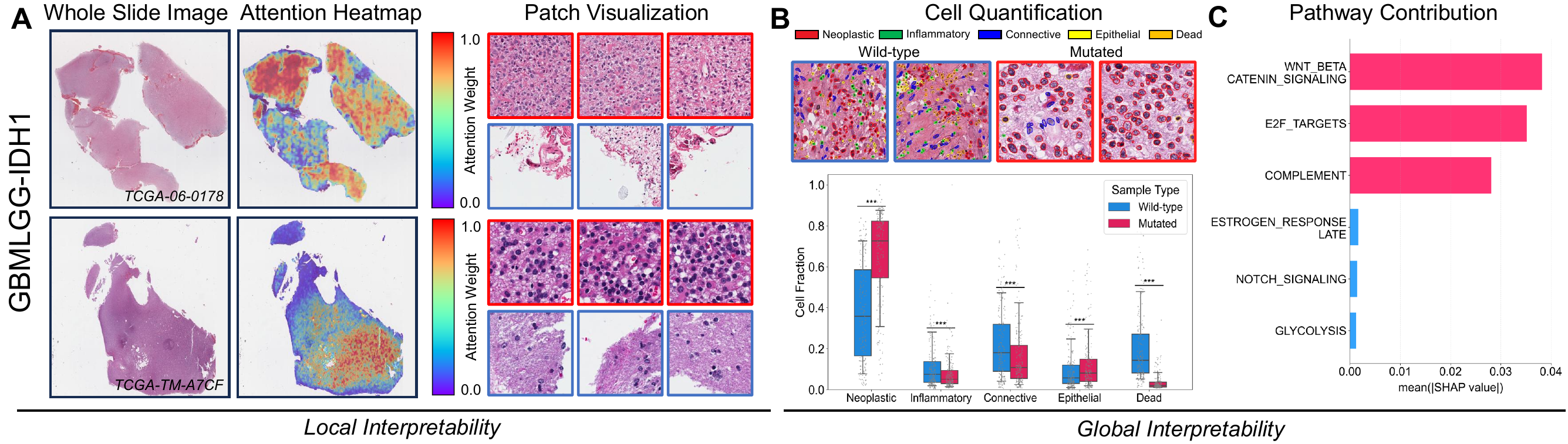}
	\caption{Local and global interpretability analyses of PathLUPI glioma cancer with \textit{IDH1} mutation status. 
		\textbf{a.} Spatial attention heatmaps highlighting histological regions most relevant for molecular prediction in mutated samples. 
		\textbf{b.} Quantification of cellular composition (neoplastic, inflammatory, epithelial, connective, dead cells) in the top 1\% high-attention patches across the cohort. 
		\textbf{c.} Shapley value attribution of the top-3 and bottom-3 transcriptomic pathways influencing model predictions in each cohort.}
	\label{fig:supp-10}
\end{figure}

\begin{figure}[h]
	\centering
	\includegraphics[width=\linewidth]{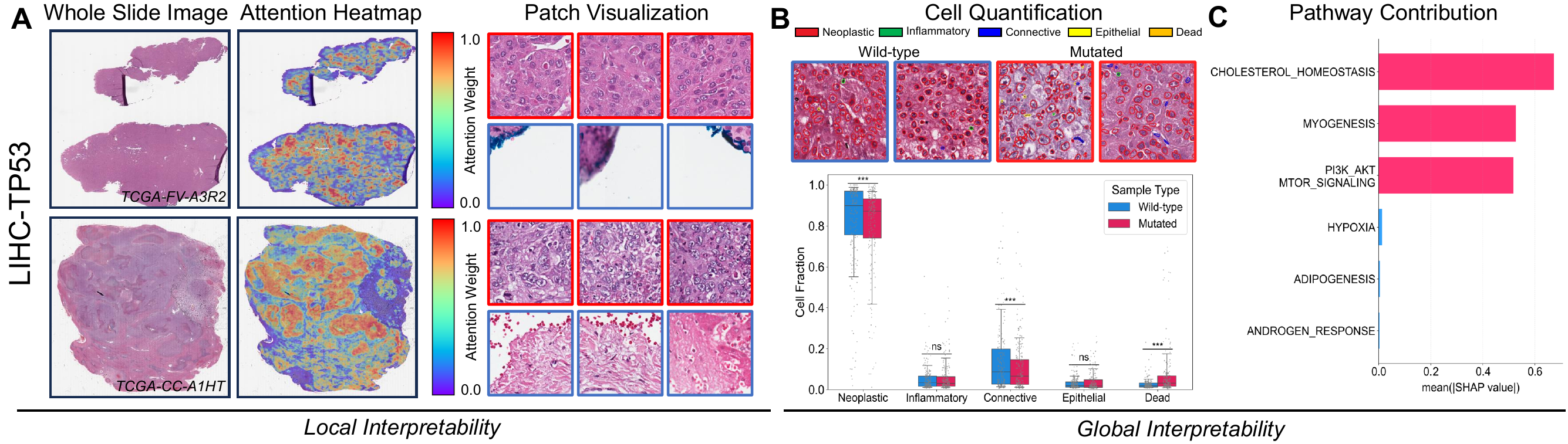}
	\caption{Local and global interpretability analyses of PathLUPI liver hepatocellular carcinoma with \textit{TP53} mutation status. 
		\textbf{a.} Spatial attention heatmaps highlighting histological regions most relevant for molecular prediction in mutated samples. 
		\textbf{b.} Quantification of cellular composition (neoplastic, inflammatory, epithelial, connective, dead cells) in the top 1\% high-attention patches across the cohort. 
		\textbf{c.} Shapley value attribution of the top-3 and bottom-3 transcriptomic pathways influencing model predictions in each cohort.}
	\label{fig:supp-11}
\end{figure}

\begin{figure}[h]
	\centering
	\includegraphics[width=\linewidth]{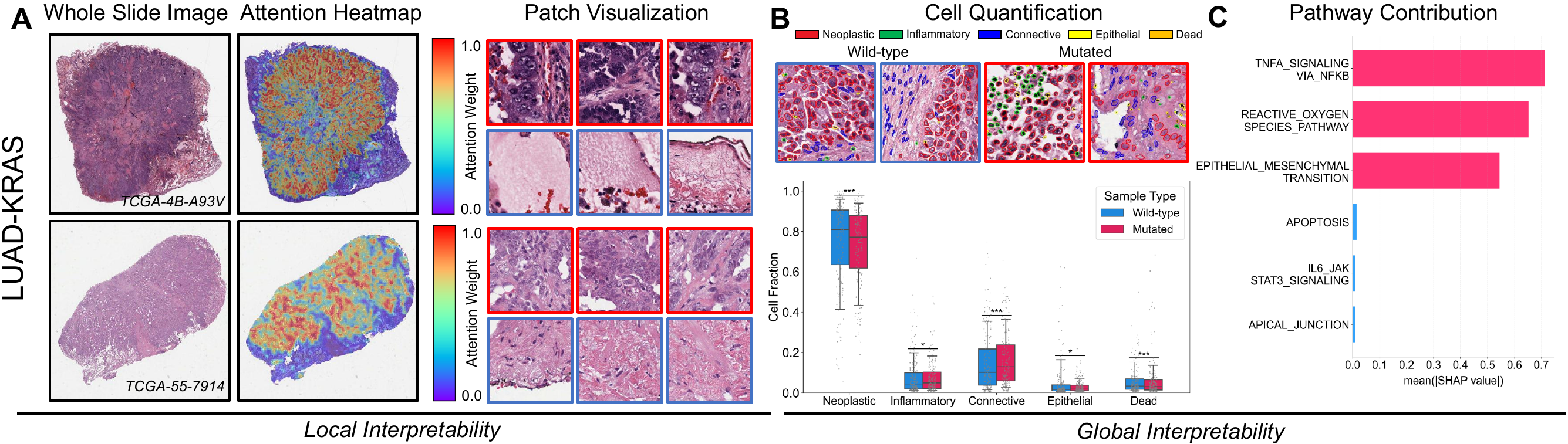}
	\caption{Local and global interpretability analyses of PathLUPI lung adenocarcinoma with \textit{KRAS} mutation status. 
		\textbf{a.} Spatial attention heatmaps highlighting histological regions most relevant for molecular prediction in mutated samples. 
		\textbf{b.} Quantification of cellular composition (neoplastic, inflammatory, epithelial, connective, dead cells) in the top 1\% high-attention patches across the cohort. 
		\textbf{c.} Shapley value attribution of the top-3 and bottom-3 transcriptomic pathways influencing model predictions in each cohort.}
	\label{fig:supp-12}
\end{figure}

\begin{figure}[h]
	\centering
	\includegraphics[width=\linewidth]{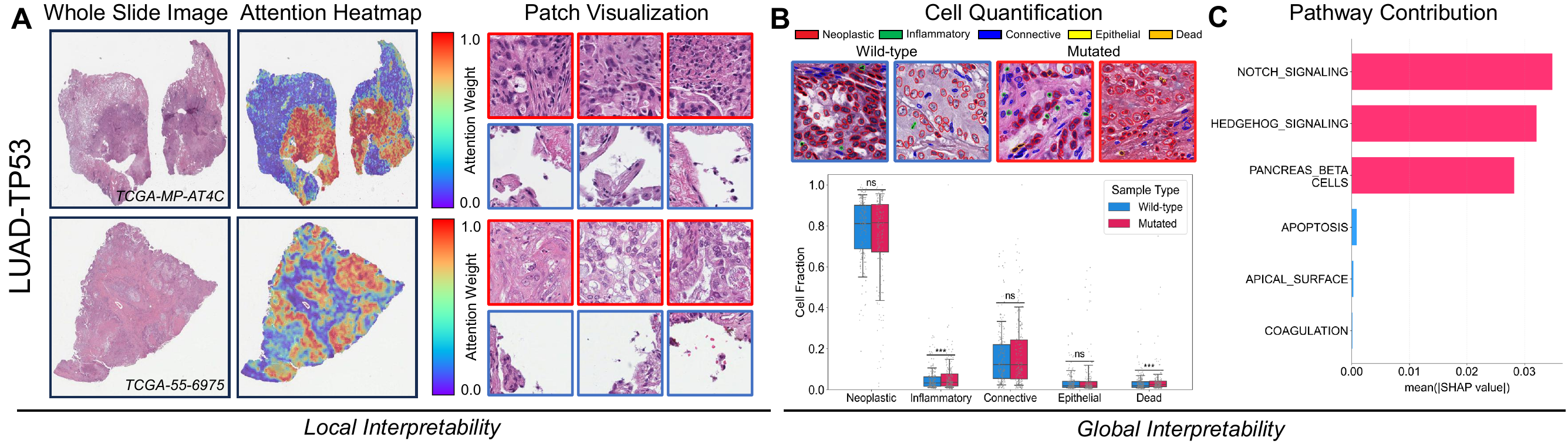}
	\caption{Local and global interpretability analyses of PathLUPI lung adenocarcinoma with \textit{TP53} mutation status. 
		\textbf{a.} Spatial attention heatmaps highlighting histological regions most relevant for molecular prediction in mutated samples. 
		\textbf{b.} Quantification of cellular composition (neoplastic, inflammatory, epithelial, connective, dead cells) in the top 1\% high-attention patches across the cohort. 
		\textbf{c.} Shapley value attribution of the top-3 and bottom-3 transcriptomic pathways influencing model predictions in each cohort.}
	\label{fig:supp-13}
\end{figure}

\begin{figure}[h]
	\centering
	\includegraphics[width=\linewidth]{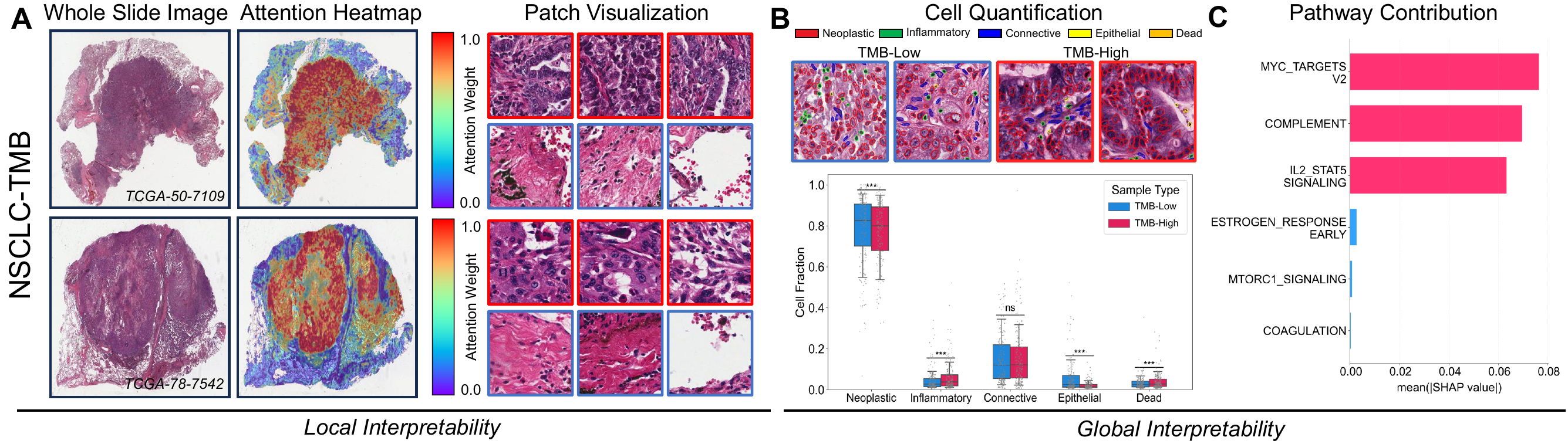}
	\caption{Local and global interpretability analyses of PathLUPI non-small-cell lung carcinoma with tumor mutational burden status. 
		\textbf{a.} Spatial attention heatmaps highlighting histological regions most relevant for molecular prediction in mutated samples. 
		\textbf{b.} Quantification of cellular composition (neoplastic, inflammatory, epithelial, connective, dead cells) in the top 1\% high-attention patches across the cohort. 
		\textbf{c.} Shapley value attribution of the top-3 and bottom-3 transcriptomic pathways influencing model predictions in each cohort.}
	\label{fig:supp-14}
\end{figure}

\begin{figure}[h]
	\centering
	\includegraphics[width=\linewidth]{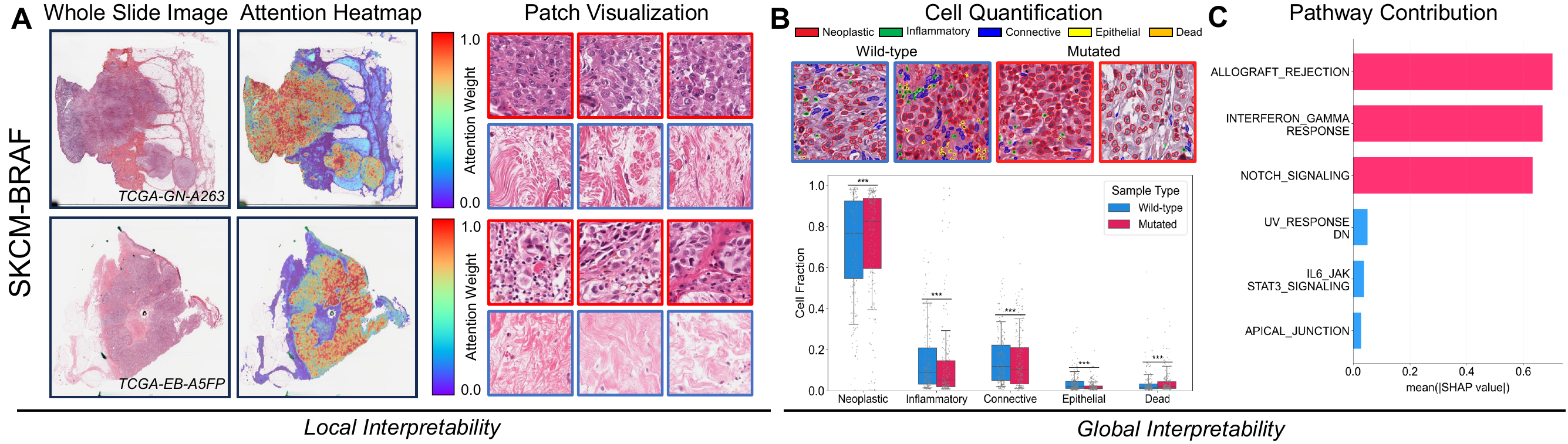}
	\caption{Local and global interpretability analyses of PathLUPI skin cutaneous melanoma with \textit{BRAF} mutation status.
		\textbf{a.} Spatial attention heatmaps highlighting histological regions most relevant for molecular prediction in mutated samples. 
		\textbf{b.} Quantification of cellular composition (neoplastic, inflammatory, epithelial, connective, dead cells) in the top 1\% high-attention patches across the cohort. 
		\textbf{c.} Shapley value attribution of the top-3 and bottom-3 transcriptomic pathways influencing model predictions in each cohort.}
	\label{fig:supp-15}
\end{figure}

\begin{table}[h]
	\fontsize{8}{10}\selectfont
	\centering
	\begin{tabular}{l|l|ccccc}
		\toprule
		Datasets & Metrics & ABMIL               & CLAM                & DTFD                & TransMIL            & PathLUPI                  \\
									\midrule
		\multirow{3}{*}{BRCA-PR}    & AUC     & 0.825 (0.824-0.827) & 0.826 (0.824-0.827) & 0.830 (0.829-0.831) & 0.814 (0.813-0.816) & \textbf{0.849 (0.848-0.850)}   \\
		                            & ACC     & 0.792 (0.791-0.794) & 0.807 (0.806-0.808) & 0.793 (0.792-0.795) & 0.737 (0.735-0.739) & \textbf{0.816 (0.815-0.817)}   \\
		                            & F1      & 0.735 (0.733-0.737) & 0.766 (0.765-0.767) & 0.732 (0.730-0.733)  & 0.702 (0.700-0.704) & \textbf{0.778 (0.777-0.779)}   \\
		\midrule
		\multirow{3}{*}{BRCA-PR*}   & AUC     & 0.738 (0.737-0.739) & 0.731 (0.730-0.731) & 0.739 (0.738-0.740) & 0.741 (0.741-0.741) & \textbf{0.753 (0.752-0.754)}   \\
		                            & ACC     & 0.676 (0.676-0.676) & 0.659 (0.658-0.659) & 0.648 (0.646-0.650) & 0.664 (0.664-0.665) & \textbf{0.701 (0.700-0.702)}   \\
		                            & F1      & 0.629 (0.627-0.631) & 0.624 (0.623-0.625) & 0.635 (0.633-0.637) & 0.613 (0.612-0.615) & \textbf{0.640 (0.637-0.643)}   \\
		\midrule
		\multirow{3}{*}{BRCA-TP53}  & AUC     & 0.858 (0.857-0.859) & 0.856 (0.855-0.857) & 0.859 (0.858-0.860) & 0.853 (0.852-0.854) & \textbf{0.878 (0.877-0.879)}   \\
		                            & ACC     & \textbf{0.816 (0.815-0.817)} & 0.808 (0.807-0.809) & 0.806 (0.805-0.808) & 0.790 (0.789-0.791) & 0.814 (0.814-0.815)   \\
		                            & F1      & \textbf{0.788 (0.787-0.789)} & 0.769 (0.768-0.771) & 0.759 (0.758-0.761) & 0.741 (0.740-0.743) & 0.774 (0.773-0.774)   \\
		\midrule
		\multirow{3}{*}{BRCA-TP53*} & AUC     & \textbf{0.785 (0.784-0.785)} & 0.745 (0.745-0.746) & 0.767 (0.766-0.768) & 0.782 (0.781-0.783) & 0.774 (0.773-0.775)   \\
		                            & ACC     & 0.705 (0.704-0.706) & 0.618 (0.617-0.620) & 0.655 (0.652-0.657) & 0.689 (0.688-0.690) & \textbf{0.723 (0.723-0.724)}   \\
		                            & F1      & 0.633 (0.632-0.634) & 0.621 (0.621-0.622) & 0.623 (0.622-0.625) & \textbf{0.657 (0.657-0.657)} & 0.612 (0.610-0.615)   \\
		\midrule
		\multirow{3}{*}{CRC-TP53}   & AUC     & 0.778 (0.776-0.780) & 0.784 (0.782-0.786) & 0.789 (0.786-0.791) & 0.699 (0.696-0.702) & \textbf{0.806 (0.803-0.808)}   \\
		                            & ACC     & \textbf{0.722 (0.720-0.724)} & 0.697 (0.695-0.698) & 0.700 (0.699-0.702) & 0.705 (0.703-0.707) & 0.721 (0.718-0.724)   \\
		                            & F1      & 0.666 (0.664-0.668) & 0.632 (0.630-0.634) & 0.609 (0.606-0.611) & 0.621 (0.618-0.625) & \textbf{0.684 (0.682-0.687)}   \\
		\midrule
		\multirow{3}{*}{LIHC-TP53}  & AUC     & 0.769 (0.767-0.771) & 0.768 (0.766-0.770) & 0.750 (0.748-0.753) & 0.752 (0.749-0.754) & \textbf{0.778 (0.776-0.780)}   \\
		                            & ACC     & \textbf{0.770 (0.768-0.771)} & 0.756 (0.754-0.757) & 0.743 (0.741-0.744) & 0.744 (0.742-0.745) & 0.724 (0.722-0.726)   \\
		                            & F1      & \textbf{0.688 (0.687-0.690)} & 0.676 (0.674-0.678) & 0.666 (0.664-0.668) & 0.627 (0.624-0.630) & 0.604 (0.601-0.608)   \\
		\midrule
		\multirow{3}{*}{LUAD-TP53}  & AUC     & 0.745 (0.743-0.747) & 0.746 (0.744-0.748) & 0.751 (0.749-0.752) & 0.717 (0.715-0.718) & \textbf{0.791 (0.789-0.792)}   \\
		                            & ACC     & 0.685 (0.684-0.687) & 0.679 (0.677-0.680) & 0.706 (0.704-0.707) & 0.629 (0.627-0.631) & \textbf{0.725 (0.723-0.726)}   \\
		                            & F1      & 0.680 (0.679-0.682) & 0.673 (0.671-0.674) & 0.701 (0.699-0.702) & 0.604 (0.602-0.607) & \textbf{0.716 (0.714-0.718)}   \\
		\midrule
		\multirow{3}{*}{NSCLC-TMB}  & AUC     & 0.694 (0.693-0.696) & 0.699 (0.697-0.701) & 0.690 (0.688-0.691) & 0.700 (0.698-0.702) & \textbf{0.735 (0.733-0.736)}   \\
		                            & ACC     & 0.756 (0.755-0.757) & 0.752 (0.750-0.753) & 0.754 (0.753-0.755) & 0.684 (0.683-0.685) & \textbf{0.761 (0.760-0.763)}   \\
		                            & F1      & 0.562 (0.560-0.565) & 0.534 (0.531-0.536) & 0.540 (0.538-0.542) & 0.536 (0.533-0.539) & \textbf{0.609 (0.608-0.611)}   \\
		\bottomrule
	\end{tabular}
        \caption{Performance of the investigational biomarker prediction task. Best-performing model for each metric is in \textbf{bold}. * denotes external cohorts.}
        \label{tab:quantitative-1}
\end{table}

\begin{table}[h]
	\fontsize{8}{10}\selectfont
	\centering
	\begin{tabular}{l|l|ccccc}
		\toprule
		Datasets    & Metrics & ABMIL               & CLAM                & DTFD                & TransMIL            & PathLUPI                \\
									  \midrule
		\multirow{3}{*}{BLCA-FGFR3}   & AUC     & 0.845 (0.843-0.848) & 0.845 (0.843-0.847) & 0.841 (0.839-0.844) & 0.796 (0.793-0.799) & \textbf{0.861 (0.859-0.864)} \\
		                              & ACC     & 0.857 (0.856-0.859) & 0.882 (0.880-0.883) & 0.855 (0.854-0.857) & 0.850 (0.849-0.852) & \textbf{0.899 (0.898-0.901)} \\
		                              & F1      & 0.626 (0.623-0.629) & 0.707 (0.703-0.710) & 0.673 (0.670-0.675) & 0.665 (0.663-0.668) & \textbf{0.766 (0.764-0.769)} \\
		\midrule
		\multirow{3}{*}{BRCA-ER}      & AUC     & 0.902 (0.901-0.903) & 0.904 (0.902-0.905) & 0.898 (0.897-0.899) & 0.904 (0.902-0.905) & \textbf{0.918 (0.917-0.919)} \\
		                              & ACC     & 0.872 (0.871-0.873) & 0.860 (0.859-0.861) & 0.875 (0.875-0.876) & 0.859 (0.858-0.860) & \textbf{0.885 (0.884-0.886)} \\
		                              & F1      & 0.801 (0.800-0.803) & 0.761 (0.759-0.763) & 0.811 (0.810-0.812) & 0.777 (0.775-0.779) & \textbf{0.824 (0.822-0.825)} \\
		\midrule
		\multirow{3}{*}{BRCA-ER*}     & AUC     & 0.695 (0.695-0.696) & 0.697 (0.696-0.697) & 0.688 (0.687-0.688) & 0.706 (0.705-0.706) & \textbf{0.736 (0.735-0.736)} \\
		                              & ACC     & \textbf{0.741 (0.741-0.741)} & 0.708 (0.706-0.709) & 0.649 (0.647-0.651) & 0.703 (0.703-0.704) & 0.727 (0.725-0.730) \\
		                              & F1      & 0.567 (0.566-0.568) & 0.520 (0.518-0.521) & 0.559 (0.558-0.559) & \textbf{0.596 (0.596-0.597)} & 0.578 (0.577-0.579) \\
		\midrule
		\multirow{3}{*}{BRCA-HER2}    & AUC     & 0.722 (0.720-0.724) & 0.721 (0.719-0.723) & 0.727 (0.725-0.730) & 0.705 (0.703-0.707) & \textbf{0.742 (0.740-0.744)} \\
		                              & ACC     & 0.776 (0.775-0.777) & 0.762 (0.760-0.764) & 0.783 (0.782-0.784) & 0.772 (0.771-0.774) & \textbf{0.785 (0.784-0.786)} \\
		                              & F1      & 0.544 (0.542-0.547) & 0.555 (0.554-0.557) & 0.543 (0.540-0.546) & 0.558 (0.556-0.560) & \textbf{0.607 (0.605-0.609)} \\
		\midrule
		\multirow{3}{*}{BRCA-HER2*}   & AUC     & 0.665 (0.664-0.665) & 0.662 (0.661-0.664) & 0.664 (0.663-0.664) & 0.583 (0.580-0.586) & \textbf{0.678 (0.676-0.680)} \\
		                              & ACC     & 0.630 (0.629-0.631) & \textbf{0.633 (0.631-0.635)} & 0.604 (0.595-0.608) & 0.588 (0.582-0.594) & 0.621 (0.619-0.622) \\
		                              & F1      & 0.559 (0.556-0.562) & 0.564 (0.561-0.567) & 0.560 (0.559-0.560) & 0.486 (0.483-0.489) & \textbf{0.601 (0.599-0.602)} \\
		\midrule
		\multirow{3}{*}{BRCA-TNBC}    & AUC     & 0.912 (0.910-0.913) & 0.903 (0.901-0.904) & 0.910 (0.909-0.911) & 0.907 (0.906-0.908) & \textbf{0.924 (0.923-0.925)} \\
		                              & ACC     & 0.889 (0.888-0.890) & 0.881 (0.880-0.882) & 0.883 (0.883-0.884) & 0.888 (0.887-0.889) & \textbf{0.895 (0.895-0.896)} \\
		                              & F1      & 0.788 (0.787-0.790) & 0.758 (0.757-0.759) & 0.759 (0.757-0.761) & 0.765 (0.762-0.768) & \textbf{0.802 (0.801-0.804)} \\
		\midrule
		\multirow{3}{*}{BRCA-TNBC*}   & AUC     & 0.759 (0.758-0.759) & 0.767 (0.766-0.767) & 0.752 (0.752-0.752) & 0.729 (0.729-0.730) & \textbf{0.792 (0.791-0.792)} \\
		                              & ACC     & 0.886 (0.885-0.887) & 0.893 (0.893-0.894) & 0.850 (0.848-0.851) & 0.887 (0.887-0.887) & \textbf{0.903 (0.901-0.905)} \\
		                              & F1      & 0.583 (0.582-0.584) & 0.573 (0.572-0.574) & 0.575 (0.574-0.575) & 0.582 (0.582-0.583) & \textbf{0.595 (0.594-0.596)} \\
		\midrule
		\multirow{3}{*}{BRCA-PIK3CA}  & AUC     & 0.694 (0.693-0.696) & 0.695 (0.694-0.696) & 0.686 (0.685-0.688) & 0.696 (0.695-0.697) & \textbf{0.726 (0.725-0.727)} \\
		                              & ACC     & 0.663 (0.662-0.664) & 0.664 (0.663-0.665) & 0.669 (0.667-0.670) & \textbf{0.689 (0.688-0.690)} & 0.685 (0.684-0.686) \\
		                              & F1      & 0.563 (0.561-0.566) & 0.549 (0.547-0.552) & 0.596 (0.595-0.598) & \textbf{0.640 (0.638-0.641)} & 0.622 (0.620-0.624) \\
		\midrule
		\multirow{3}{*}{BRCA-PIK3CA*} & AUC     & 0.620 (0.620-0.621) & \textbf{0.632 (0.631-0.632)} & 0.585 (0.583-0.586) & 0.625 (0.624-0.626) & 0.631 (0.628-0.633) \\
		                              & ACC     & 0.463 (0.462-0.463) & 0.430 (0.428-0.433) & \textbf{0.548 (0.545-0.550)} & 0.444 (0.442-0.447) & 0.535 (0.532-0.539) \\
		                              & F1      & 0.510 (0.509-0.511) & 0.430 (0.428-0.433) & 0.366 (0.362-0.369) & 0.518 (0.518-0.519) & \textbf{0.519 (0.517-0.520)} \\
		\midrule
		\multirow{3}{*}{CRC-BRAF}     & AUC     & 0.737 (0.734-0.740) & 0.727 (0.724-0.730) & 0.762 (0.759-0.765) & 0.690 (0.686-0.694) & \textbf{0.840 (0.837-0.843)} \\
		                              & ACC     & 0.843 (0.842-0.845) & 0.847 (0.846-0.849) & \textbf{0.869 (0.867-0.871)} & 0.852 (0.850-0.853) & 0.856 (0.854-0.858) \\
		                              & F1      & 0.525 (0.522-0.527) & 0.549 (0.545-0.552) & 0.552 (0.549-0.556) & 0.560 (0.557-0.564) & \textbf{0.591 (0.587-0.595)} \\
		\midrule
		\multirow{3}{*}{CRC-BRAF*}    & AUC     & 0.610 (0.609-0.611) & 0.626 (0.626-0.627) & 0.564 (0.562-0.566) & 0.590 (0.590-0.591) & \textbf{0.655 (0.653-0.657)} \\
		                              & ACC     & 0.643 (0.639-0.647) & 0.635 (0.630-0.641) & 0.737 (0.735-0.738) & 0.663 (0.659-0.666) & \textbf{0.755 (0.752-0.758)} \\
		                              & F1      & 0.347 (0.346-0.348) & 0.357 (0.355-0.359) & 0.238 (0.236-0.241) & 0.359 (0.358-0.360) & \textbf{0.442 (0.440-0.445)} \\
		\midrule
		\multirow{3}{*}{CRC-KRAS}     & AUC     & 0.617 (0.614-0.620) & 0.641 (0.638-0.643) & 0.630 (0.627-0.633) & 0.601 (0.599-0.604) & \textbf{0.705 (0.703-0.708)} \\
		                              & ACC     & 0.578 (0.576-0.581) & 0.577 (0.575-0.580) & \textbf{0.623 (0.621-0.625)} & 0.519 (0.516-0.523) & 0.583 (0.580-0.587) \\
		                              & F1      & 0.527 (0.523-0.530) & 0.561 (0.559-0.564) & 0.553 (0.549-0.556) & 0.463 (0.459-0.467) & \textbf{0.565 (0.561-0.569)} \\
		\midrule
		\multirow{3}{*}{GBMLGG-IDH1}  & AUC     & 0.984 (0.983-0.984) & 0.982 (0.982-0.983) & 0.981 (0.980-0.981) & 0.986 (0.985-0.986) & \textbf{0.992 (0.992-0.992)} \\
		                              & ACC     & 0.928 (0.927-0.929) & 0.933 (0.932-0.934) & \textbf{0.939 (0.938-0.940)} & 0.921 (0.920-0.922) & 0.923 (0.921-0.924) \\
		                              & F1      & 0.926 (0.925-0.927) & 0.932 (0.930-0.933) & \textbf{0.937 (0.936-0.939)} & 0.918 (0.917-0.919) & 0.921 (0.919-0.923) \\
		\midrule
		\multirow{3}{*}{GBMLGG-IDH1*} & AUC     & 0.912 (0.912-0.912) & 0.915 (0.914-0.915) & 0.914 (0.913-0.914) & 0.914 (0.914-0.914) & \textbf{0.920 (0.919-0.920)} \\
		                              & ACC     & \textbf{0.852 (0.851-0.852)} & 0.836 (0.836-0.837) & 0.842 (0.841-0.842) & 0.824 (0.823-0.824) & 0.850 (0.849-0.850) \\
		                              & F1      & \textbf{0.817 (0.816-0.817)} & 0.797 (0.796-0.798) & 0.793 (0.792-0.794) & 0.792 (0.792-0.792) & 0.804 (0.804-0.805) \\
		\midrule
		\multirow{3}{*}{LUAD-KRAS}    & AUC     & 0.598 (0.596-0.600) & 0.577 (0.574-0.579) & 0.619 (0.617-0.621) & 0.635 (0.632-0.637) & \textbf{0.671 (0.669-0.673)} \\
		                              & ACC     & 0.698 (0.696-0.700) & \textbf{0.704 (0.702-0.705)} & 0.681 (0.680-0.683) & 0.628 (0.626-0.630) & 0.672 (0.670-0.674) \\
		                              & F1      & 0.471 (0.469-0.473) & 0.491 (0.489-0.493) & \textbf{0.513 (0.510-0.515)} & 0.510 (0.508-0.512) & 0.509 (0.507-0.512) \\
		\midrule
		\multirow{3}{*}{LUAD-EGFR}    & AUC     & 0.700 (0.697-0.703) & 0.702 (0.698-0.705) & 0.733 (0.730-0.736) & 0.707 (0.704-0.710) & \textbf{0.769 (0.766-0.772)} \\
		                              & ACC     & 0.872 (0.871-0.874) & \textbf{0.874 (0.873-0.875)} & 0.864 (0.862-0.865) & 0.872 (0.871-0.873) & 0.872 (0.870-0.873) \\
		                              & F1      & 0.466 (0.465-0.466) & 0.466 (0.466-0.466) & \textbf{0.517 (0.515-0.520)} & 0.495 (0.493-0.497) & 0.492 (0.491-0.494) \\
		\midrule
		\multirow{3}{*}{SKCM-BRAF}    & AUC     & 0.603 (0.601-0.605) & 0.604 (0.602-0.606) & 0.583 (0.581-0.586) & 0.631 (0.629-0.633) & \textbf{0.684 (0.682-0.686)} \\
		                              & ACC     & 0.602 (0.600-0.604) & 0.592 (0.590-0.593) & 0.598 (0.596-0.600) & 0.574 (0.571-0.577) & \textbf{0.625 (0.623-0.627)} \\
		                              & F1      & 0.565 (0.563-0.567) & 0.575 (0.573-0.577) & 0.576 (0.573-0.578) & 0.557 (0.554-0.561) & \textbf{0.614 (0.613-0.616)} \\
		\bottomrule
	\end{tabular}
    \caption{Performance of actionable biomarker prediction. Best-performing model for each metric is in \textbf{bold}. * denotes external cohorts.}
    \label{tab:quantitative-2}
\end{table}

\begin{table}[h]
	\fontsize{8}{10}\selectfont
	\centering
	\begin{tabular}{l|l|ccccc}
		\toprule
		Datasets  & Metrics & ABMIL               & CLAM                & DTFD                & TransMIL            & PathLUPI                \\
								\midrule
		\multirow{3}{*}{BLCA}   & AUC     & 0.887 (0.886-0.888) & 0.887 (0.886-0.888) & 0.884 (0.884-0.885) & 0.881 (0.880-0.883) & \textbf{0.900 (0.899-0.901)} \\
		                        & ACC     & 0.697 (0.695-0.699) & 0.694 (0.692-0.696) & 0.707 (0.705-0.709) & \textbf{0.722 (0.720-0.724)} & 0.717 (0.715-0.719) \\
		                        & F1      & 0.601 (0.599-0.603) & 0.568 (0.566-0.571) & 0.600 (0.598-0.603) & \textbf{0.624 (0.621-0.626)} & 0.615 (0.613-0.618) \\
								\midrule
								\multirow{3}{*}{BRCA}   & AUC     & 0.867 (0.866-0.868) & 0.867 (0.866-0.868) & 0.858 (0.858-0.859) & 0.852 (0.851-0.853) & \textbf{0.876 (0.876-0.877)} \\
		                        & ACC     & \textbf{0.689 (0.688-0.690)} & \textbf{0.689 (0.688-0.690)} & 0.664 (0.663-0.665) & 0.642 (0.641-0.643) & 0.676 (0.675-0.677) \\
		                        & F1      & \textbf{0.633 (0.631-0.635)} & 0.614 (0.612-0.616) & 0.602 (0.600-0.603) & 0.516 (0.515-0.518) & 0.606 (0.605-0.608) \\
								\midrule
								\multirow{3}{*}{BRCA*}  & AUC     & 0.701 (0.701-0.702) & 0.706 (0.705-0.706) & 0.666 (0.666-0.667) & 0.710 (0.710-0.710) & \textbf{0.727 (0.726-0.727)} \\
		                        & ACC     & 0.425 (0.424-0.426) & 0.425 (0.425-0.426) & 0.354 (0.353-0.355) & 0.485 (0.485-0.486) & \textbf{0.491 (0.490-0.493)} \\
		                        & F1      & 0.355 (0.354-0.356) & 0.357 (0.356-0.357) & 0.260 (0.259-0.261) & 0.416 (0.416-0.417) & \textbf{0.450 (0.449-0.452)} \\
								\midrule
								\multirow{3}{*}{CRC}    & AUC     & 0.822 (0.821-0.823) & 0.818 (0.817-0.819) & 0.810 (0.809-0.811) & 0.793 (0.791-0.794) & \textbf{0.831 (0.830-0.832)} \\
		                        & ACC     & 0.591 (0.589-0.593) & \textbf{0.628 (0.627-0.630)} & 0.605 (0.603-0.607) & 0.567 (0.565-0.569) & 0.614 (0.612-0.616) \\
		                        & F1      & 0.531 (0.528-0.533) & \textbf{0.568 (0.566-0.570)} & 0.547 (0.545-0.550) & 0.502 (0.500-0.504) & 0.556 (0.554-0.557) \\
								\midrule
								\multirow{3}{*}{GBMLGG} & AUC     & 0.867 (0.866-0.868) & 0.865 (0.864-0.866) & 0.861 (0.860-0.862) & 0.870 (0.869-0.871) & \textbf{0.886 (0.885-0.887)} \\
		                        & ACC     & 0.684 (0.682-0.685) & \textbf{0.691 (0.689-0.693)} & 0.674 (0.673-0.676) & 0.685 (0.683-0.686) & 0.684 (0.683-0.686) \\
		                        & F1      & 0.510 (0.508-0.512) & 0.523 (0.521-0.526) & 0.506 (0.504-0.508) & 0.511 (0.509-0.513) & \textbf{0.525 (0.523-0.527)} \\
								\midrule
								\multirow{3}{*}{HNSC}   & AUC     & 0.775 (0.773-0.776) & 0.779 (0.778-0.781) & 0.757 (0.755-0.759) & 0.760 (0.758-0.762) & \textbf{0.803 (0.802-0.804)} \\
		                        & ACC     & 0.475 (0.472-0.479) & 0.498 (0.494-0.501) & 0.429 (0.427-0.432) & 0.542 (0.539-0.544) & \textbf{0.565 (0.563-0.568)} \\
		                        & F1      & 0.436 (0.433-0.439) & 0.446 (0.444-0.449) & 0.403 (0.401-0.406) & 0.486 (0.484-0.489) & \textbf{0.519 (0.516-0.521)} \\
								\midrule
								\multirow{3}{*}{PanGI}  & AUC     & 0.861 (0.860-0.862) & 0.863 (0.862-0.863) & 0.853 (0.852-0.854) & 0.832 (0.831-0.833) & \textbf{0.873 (0.873-0.874)} \\
		                        & ACC     & 0.739 (0.738-0.741) & 0.490 (0.489-0.491) & 0.422 (0.420-0.424) & 0.534 (0.533-0.535) & \textbf{0.753 (0.752-0.754)} \\
		                        & F1      & 0.541 (0.539-0.543) & 0.540 (0.538-0.543) & 0.519 (0.516-0.522) & 0.456 (0.453-0.458) & \textbf{0.547 (0.545-0.549)} \\
								\midrule
								\multirow{3}{*}{UCEC}   & AUC     & 0.808 (0.807-0.809) & 0.811 (0.810-0.812) & 0.799 (0.798-0.800) & 0.787 (0.786-0.788) & \textbf{0.822 (0.821-0.823)} \\
		                        & ACC     & 0.579 (0.577-0.582) & 0.584 (0.582-0.586) & 0.576 (0.575-0.578) & 0.580 (0.578-0.582) & \textbf{0.595 (0.594-0.597)} \\
		                        & F1      & 0.494 (0.491-0.496) & 0.502 (0.500-0.503) & 0.511 (0.509-0.512) & 0.502 (0.500-0.503) & \textbf{0.533 (0.531-0.535)} \\
		\bottomrule
	\end{tabular}
    \caption{Performance of molecular subtyping. Best-performing model for each metric is in \textbf{bold}.  * denotes external cohorts.}
    \label{tab:quantitative-3}
\end{table}

\begin{table}[h]
	\fontsize{8}{10}\selectfont
	\centering
	\begin{tabular}{l|l|ccccc}
		\toprule
		    Datasets   & Metrics & ABMIL               & CLAM                & DTFD                & TransMIL            & PathLUPI                \\
			   \midrule
		BLCA   & C-index & 0.633 (0.631-0.635) & 0.631 (0.629-0.633) & 0.622 (0.620-0.624) & 0.640 (0.638-0.641) & \textbf{0.651 (0.649-0.653)} \\
		BRCA   & C-index & 0.668 (0.666-0.670) & 0.668 (0.666-0.670) & 0.657 (0.655-0.659) & 0.677 (0.674-0.679) & \textbf{0.710 (0.708-0.712)} \\
		BRCA*  & C-index & 0.625 (0.624-0.625) & 0.622 (0.621-0.622) & 0.587 (0.586-0.589) & 0.614 (0.614-0.615) & \textbf{0.641 (0.640-0.641)} \\
		CRC    & C-index & 0.683 (0.681-0.685) & 0.685 (0.683-0.687) & 0.637 (0.635-0.639) & 0.664 (0.662-0.666) & \textbf{0.692 (0.690-0.694)} \\
        GBM & C-index & 0.512 (0.507-0.515) & 0.575 (0.572-0.580) & 0.544 (0.540-0.546) & 0.537 (0.534-0.541) & \textbf{0.662 (0.658-0.665)} \\
        LGG & C-index & 0.755 (0.752-0.757) & 0.753 (0.750-0.756) & 0.751 (0.749-0.754) & 0.766 (0.763-0.768) & \textbf{0.786 (0.784-0.788)} \\
		HNSC   & C-index & 0.607 (0.606-0.609) & 0.608 (0.607-0.610) & 0.609 (0.607-0.610) & 0.626 (0.625-0.628) & \textbf{0.631 (0.629-0.632)} \\
		KIRC   & C-index & 0.732 (0.730-0.733) & 0.732 (0.731-0.734) & 0.733 (0.731-0.735) & 0.728 (0.726-0.729) & \textbf{0.747 (0.746-0.749)} \\
		LIHC   & C-index & 0.695 (0.693-0.698) & 0.699 (0.696-0.702) & 0.689 (0.686-0.692) & 0.716 (0.713-0.718) & \textbf{0.776 (0.773-0.778)} \\
		LUAD   & C-index & 0.619 (0.617-0.621) & 0.621 (0.619-0.623) & 0.612 (0.610-0.615) & 0.643 (0.641-0.645) & \textbf{0.668 (0.667-0.670)} \\
		LUSC   & C-index & 0.553 (0.551-0.555) & 0.548 (0.546-0.549) & 0.575 (0.573-0.577) & 0.590 (0.588-0.592) & \textbf{0.618 (0.616-0.620)} \\
		LUSC*  & C-index & 0.605 (0.604-0.607) & 0.624 (0.623-0.625) & 0.631 (0.630-0.632) & 0.621 (0.620-0.621) & \textbf{0.649 (0.647-0.650)} \\
		SKCM   & C-index & 0.623 (0.621-0.625) & 0.618 (0.617-0.620) & 0.641 (0.639-0.642) & 0.627 (0.626-0.629) & \textbf{0.651 (0.649-0.653)} \\
		STAD   & C-index & 0.629 (0.626-0.631) & 0.640 (0.638-0.642) & 0.648 (0.646-0.650) & 0.634 (0.632-0.637) & \textbf{0.665 (0.663-0.667)} \\
		UCEC   & C-index & 0.742 (0.740-0.744) & 0.731 (0.728-0.733) & 0.742 (0.740-0.744) & 0.720 (0.718-0.722) & \textbf{0.756 (0.754-0.758)} \\
		UCEC*  & C-index & 0.583 (0.582-0.583) & 0.594 (0.593-0.595) & 0.618 (0.617-0.619) & 0.553 (0.551-0.555) & \textbf{0.630 (0.629-0.632)} \\
		\bottomrule
	\end{tabular}
    \caption{Performance of survival prognosis. Best-performing model for each metric is in \textbf{bold}.  * denotes external cohorts.}
    \label{tab:quantitative-4}
\end{table}

\begin{table}[htbp]
	\scriptsize
	\setlength{\tabcolsep}{3.5pt}
	\renewcommand{\arraystretch}{0.9}
	\begin{tabularx}{\linewidth}{
			C{2cm}       
			C{3cm}       
			C{2.7cm}       
			C{4.5cm}       
			C{2.8cm}}             
		\toprule
		\textbf{Task Type} & \makecell{\textbf{Task ID}\\\textbf{(Patient Count)}} & \textbf{Label Type} &
		\textbf{Label Name} & \textbf{Label Source} \\ 
		\midrule
		\multirow{17}{*}{\makecell[c]{\textbf{Actionable}\\\textbf{Biomarker}\\\textbf{Prediction}}}
		& BLCA-FGFR3 (314)   & Somatic Mutation   & FGFR3 mut vs.\ wt           & TCGA (UCSC Xena)\\
		& BRCA-ER (949)      & Protein Expression & ER$+$ vs.\ ER$-$            & TCGA (UCSC Xena)\\
		& BRCA-ER* (1,527)     & Protein Expression & ER$+$ vs.\ ER$-$            & Center-1 \\
		& BRCA-HER2 (646)    & Protein Expression & HER2$+$ vs.\ HER2$-$        & TCGA (UCSC Xena)\\
		& BRCA-HER2* (1,391)    & Protein Expression & HER2$+$ vs.\ HER2$-$        & Center-1 \\
		& BRCA-PIK3CA (634)  & Somatic Mutation   & PIK3CA mut vs.\ wt          & TCGA (UCSC Xena)\\
		& BRCA-PIK3CA* (116) & Somatic Mutation   & PIK3CA mut vs.\ wt.          & CPTAC (cBioPortal)\\
		& BRCA-TNBC (976)    & Cancer Subtype & TNBC+ vs.\ TNBC-            & TCGA (UCSC Xena)\\
		& BRCA-TNBC* (1,391)   & Cancer Subtype & TNBC+ vs.\ TNBC-            & Center-1 \\
		& CRC-KRAS (333)     & Somatic Mutation   & KRAS mut. vs.\ wt.            & TCGA (UCSC Xena)\\
		& CRC-BRAF (333)     & Somatic Mutation   & BRAF mut. vs.\ wt.            & TCGA (UCSC Xena)\\
		& CRC-BRAF* (102)    & Somatic Mutation   & BRAF mut. vs.\ wt.            & CPTAC (cBioPortal)\\
		& GBMLGG-IDH1 (373)  & Somatic Mutation   & IDH1 mut. vs.\ wt.            & TCGA (UCSC Xena)\\
		& GBMLGG-IDH1* (852) & Somatic Mutation   & IDH1 mut. vs.\ wt.            & EBRAINS \\
		& LUAD-EGFR (448)    & Somatic Mutation   & EGFR mut. vs.\ wt.           & TCGA (UCSC Xena)\\
		& LUAD-KRAS (448)    & Somatic Mutation   & KRAS mut. vs.\ wt.            & TCGA (UCSC Xena)\\
		& SKCM-BRAF (412)    & Somatic Mutation   & BRAF mut. vs.\ wt.            & TCGA (UCSC Xena)\\
		\midrule
		\multirow{8}{*}{\makecell[c]{\textbf{Investigational}\\\textbf{Biomarker}\\\textbf{Prediction}}}
		& BRCA-PR (948)  & Protein Expression & PR$+$ vs.\ PR$-$            & TCGA (UCSC Xena)\\
		& BRCA-PR* (1,527)     & Protein Expression & PR$+$ vs.\ PR$-$            & Center-1\\
		& BRCA-TP53 (634)     & Somatic Mutation   & TP53 mut. vs.\ wt.            & TCGA (UCSC Xena)\\
		& BRCA-TP53* (116)   & Somatic Mutation   & TP53 mut. vs.\ wt.            & CPTAC (cBioPortal)\\
		& CRC-TP53 (333)      & Somatic Mutation   & TP53 mut. vs.\ wt.            & TCGA (UCSC Xena)\\
		& LIHC-TP53 (316)    & Somatic Mutation   & TP53 mut. vs.\ wt.            & TCGA (UCSC Xena)\\
		& LUAD-TP53 (439)    & Somatic Mutation   & TP53 mut. vs.\ wt.            & TCGA (UCSC Xena)\\
		& NSCLC-TMB (902)    & Tumor Mutation Burden & High vs.\ Low           & TCGA (UCSC Xena)\\
		\midrule
		\multirow{12}{*}{\makecell[c]{\textbf{Molecular}\\\textbf{Subtyping}}}
		& \multirow{2}{*}{BLCA-Mol (298)}      & 	\multirow{2}{*}{Molecular Subtype}  & Luminal, Luminal infiltrated, Luminal papillary, Basal squamous & \multirow{2}{*}{TCGA (TCGAbiolinks)} \\
		& \multirow{2}{*}{BRCA-Mol (505)}      & \multirow{2}{*}{Molecular Subtype}  & Luminal A, Luminal B, HER2-enriched,   & \multirow{2}{*}{TCGA (TCGAbiolinks)} \\
		&     &  &  Basal-like    & \\
		& \multirow{2}{*}{BRCA-Mol* (2,045)}      & \multirow{2}{*}{Molecular Subtype}  & Luminal A, Luminal B, HER2-enriched,   & \multirow{2}{*}{Center-2} \\
		&     &  &  Basal-like    & \\
		& CRC-Mol (492)       & Molecular Subtype  & CMS1, CMS2, CMS3, CMS4   & TCGA (TCGAbiolinks) \\
		& \multirow{2}{*}{GBMLGG-Mol (552)}    & \multirow{2}{*}{Molecular Subtype}  & G-CIMP-high, G-CIMP-low, Codel,  & \multirow{2}{*}{TCGA (TCGAbiolinks)} \\
		&     &  &  Mesenchymal-like, Classic-like    & \\
		& HNSC-Mol (218)     & Molecular Subtype  & Atypical, Basal, Classical, Mesenchymal  & TCGA (TCGAbiolinks) \\
		& PanGI-Mol (786)    & Molecular Subtype  & MSI, CIN, EBV, HM-SNV, GS   & TCGA (TCGAbiolinks) \\
		& UCEC-Mol (458)     & Molecular Subtype  & CN-high, CN-low, MSI, POLE  & TCGA (TCGAbiolinks) \\
		\midrule
		\multirow{15}{*}{\makecell[c]{\textbf{Survival}\\\textbf{Prognosis}}}
		& BLCA-Surv (376)     & Overall Survival   & OS time, Event status       & TCGA (cBioPortal)\\
		& BRCA-Surv (1,023)     & Overall Survival   & OS time, Event status       & TCGA (cBioPortal)\\
		& BRCA-Surv* (454)  & Overall Survival   & OS time, Event status       & Center-2 \\
		& CRC-Surv (579)      & Overall Survival   & OS time, Event status       & TCGA (cBioPortal)\\
		& GBM-Surv (372)  & Overall Survival   & OS time, Event status       & TCGA (cBioPortal)\\
		& HNSC-Surv (441)    & Overall Survival   & OS time, Event status       & TCGA (cBioPortal)\\
		& KIRC-Surv (498)   & Overall Survival   & OS time, Event status       & TCGA (cBioPortal)\\
        & LGG-Surv (462)  & Overall Survival   & OS time, Event status       & TCGA (cBioPortal)\\
		& LIHC-Surv (347)    & Overall Survival   & OS time, Event status       & TCGA (cBioPortal)\\
		& LUAD-Surv (455)    & Overall Survival   & OS time, Event status       & TCGA (cBioPortal)\\
		& LUSC-Surv (452)     & Overall Survival   & OS time, Event status       & TCGA (cBioPortal)\\
		& LUSC-Surv* (94)   & Overall Survival   & OS time, Event status       & CPTAC (cBioPortal)\\
		& SKCM-Surv (415)     & Overall Survival   & OS time, Event status       & TCGA (cBioPortal)\\
		& STAD-Surv (363)     & Overall Survival   & OS time, Event status       & TCGA (cBioPortal)\\
		& UCEC-Surv (495)    & Overall Survival   & OS time, Event status       & TCGA (cBioPortal)\\
		& UCEC-Surv* (94)   & Overall Survival   & OS time, Event status       & CPTAC (cBioPortal)\\
		\bottomrule
	\end{tabularx}
	\caption{Curated molecular oncology tasks grouped by task type, including survival prognosis, molecular subtyping, and biomarker prediction across multiple cancer types and data sources.}
	\label{tab:supp_tasks}
\end{table}

\begin{table}[h]
	\centering
	\small
	\begin{tabularx}{\linewidth}{p{4cm}|X}
		\toprule
		\textbf{Dataset} & \textbf{Link} \\
		\midrule
		TCGA \cite{weinstein2013cancer} & \url{https://portal.gdc.cancer.gov/} \\
		CPTAC \cite{ellis2013connecting} & \url{https://proteomic.datacommons.cancer.gov/pdc/} \\
		EBRAINS \cite{roetzer2022digital} & \url{https://www.ebrains.eu/} \\
		\bottomrule
	\end{tabularx}
	\caption{Public datasets used in this study. Some datasets may require permission before downloading.}
	\label{tab:datasets}
\end{table}

\begin{table}[h]
	\centering
	\footnotesize
	\begin{tabular}{
			>{\raggedright\arraybackslash}p{3cm} |
			>{\raggedright\arraybackslash}p{3.5cm} |
			>{\raggedright\arraybackslash}p{3.5cm} |
			>{\raggedright\arraybackslash}p{5.5cm}
		}
		\toprule
		\textbf{Biomarker} & \textbf{Targeted Therapy} & \textbf{Indication} & \textbf{Evidence} \\
		\midrule
		BLCA-FGFR3     & Erdafitinib & Bladder cancer & \url{https://www.cancer.gov/about-cancer/treatment/drugs/erdafitinib} \\
		BRCA-ER        & Tamoxifen & Breast cancer & \url{https://www.cancer.gov/about-cancer/treatment/drugs/tamoxifen} \\
		BRCA-HER2      & Trastuzumab (Herceptin) & Breast cancer & \url{https://www.cancer.gov/about-cancer/treatment/drugs/trastuzumab} \\
		BRCA-PIK3CA    & Alpelisib & Breast cancer & \url{https://www.cancer.gov/about-cancer/treatment/drugs/alpelisib} \\
		CRC-BRAF       & Encorafenib & Colorectal cancer & \url{https://www.cancer.gov/about-cancer/treatment/drugs/encorafenib} \\
		CRC-KRAS       & Sotorasib & Colorectal cancer & \url{https://www.cancer.gov/about-cancer/treatment/drugs/sotorasib} \\
		GBMLGG-IDH1    & Ivosidenib & Glioma & \url{https://www.cancer.gov/about-cancer/treatment/drugs/ivosidenib} \\
		LUAD-EGFR      & Osimertinib & Lung adenocarcinoma & \url{https://www.cancer.gov/about-cancer/treatment/drugs/osimertinib} \\
		LUAD-KRAS      & Sotorasib, Adagrasib & Lung adenocarcinoma & \url{https://www.cancer.gov/about-cancer/treatment/drugs/sotorasib}, \url{https://www.cancer.gov/about-cancer/treatment/drugs/adagrasib} \\
		SKCM-BRAF      & Vemurafenib & Melanoma & \url{https://www.cancer.gov/about-cancer/treatment/drugs/vemurafenib} \\
		\bottomrule
	\end{tabular}
\caption{Summary of clinically approved targeted therapies corresponding to actionable cancer biomarkers.}
	\label{tab:targeted_biomarkers}
\end{table}

\begin{table}[htbp]
    \centering
    \begin{tabularx}{\linewidth}{l *{4}{>{\centering\arraybackslash}X} | >{\centering\arraybackslash}X >{\centering\arraybackslash}X}
        \toprule
        \textbf{Hyperparam.} & \textbf{ABMIL} & \textbf{CLAM} & \textbf{DTFD} & \textbf{TransMIL} & \textbf{PathLUPI (WSI)} & \textbf{PathLUPI (RNASeq)} \\
        \midrule
        Input dim         & 512   & 512   & 512 & 512 & 512 & $50 \times D$ \\
        Hidden dim        & 512   & 512   & 512 & 512 & 256 & 50 $\times$ 256 \\
        Dropout           & 0.25  & 0.25  & 0.25  & 0.25 & 0.25 & 0.25 \\
        Feature dim after fusion & - & - & - & - & \multicolumn{2}{c}{$256 \times 2 \rightarrow 256$} \\
        \midrule
        \multicolumn{7}{l}{\textbf{Training Settings}} \\
        \midrule
        Batch size        & \multicolumn{6}{c}{1} \\
        Epochs            & \multicolumn{6}{c}{30} \\
        Optimizer         & \multicolumn{6}{c}{Adam}  \\
        Learning rate     & \multicolumn{6}{c}{2e-4}  \\
        Scheduler         & \multicolumn{6}{c}{Cosine}  \\
        Weight decay      & \multicolumn{6}{c}{1e-5} \\
        \bottomrule
    \end{tabularx}
    \caption{Hyperparameters for all compared models. $D$ denotes the input dimension of each pathway-grouped transcriptomic sequence.}
    \label{tab:all_hyperparams}
\end{table}